\pdfoutput=1
\documentclass[11pt, a4paper, logo]{lumia}

\usepackage[numbers,sort&compress]{natbib}
\definecolor{infrecred}{HTML}{D2071E}
\definecolor{abstrabg}{HTML}{FCDDDC}
\colorlet{abstradark}{infrecred}
\definecolor{reportaccent}{HTML}{BB0819}
\colorlet{reporttoc}{reportaccent}
\definecolor{cayenne}{HTML}{941100}
\hypersetup{allcolors=cayenne}
\definecolor{darkblue}{rgb}{0, 0, 0.5}
\definecolor{lightblue}{HTML}{D0CFEA}

\usepackage{graphicx}
\usepackage{wrapfig}
\makeatletter
\newcount\WF@toplines
\def\WF@wraphand{%
 \ifnum\c@WF@wrappedlines<\tw@ \WF@finale
 \else \begingroup
  \@tempcnta\@ne \let\WF@wrapil\relax \let\WF@noil\relax \gdef\WF@ps{}%
  \@whilenum \@tempcnta<\c@WF@wrappedlines\do{%
   \ifnum\@tempcnta>\WF@toplines \xdef\WF@ps{\WF@ps\WF@wrapil}%
   \else \xdef\WF@ps{\WF@ps\WF@noil}\fi
   \advance\@tempcnta\@ne}%
  \endgroup
  \ifx\WF@pspars\@empty \@@parshape\c@WF@wrappedlines \WF@ps \WF@noil
  \else \WF@modps \fi
 \fi}
\def\WF@mypar{\relax
 \WF@@par
 \ifnum\@@parshape=\z@ \let\WF@pspars\@empty \fi
 \global\advance\c@WF@wrappedlines-\prevgraf
 \global\advance\WF@toplines-\prevgraf
 \ifnum\WF@toplines<\z@ \global\WF@toplines\z@ \fi
 \ifnum\prevgraf>\z@ \wrapminiparbreak \fi
 \prevgraf\z@
 \ifnum\c@WF@wrappedlines<\tw@ \WF@finale \fi}
\g@addto@macro\WF@startwrapping{\ifnum\c@WF@wrappedlines>\@ne \WF@restoretol\fi}
\newdimen\wrapminislack \wrapminislack=-\maxdimen
\def\wrapminiparbreak{\global\advance\wrapminislack\parskip
  \ifdim\wrapminislack<\dimexpr\baselineskip+4pt\relax\else
    \global\advance\c@WF@wrappedlines\m@ne \global\advance\wrapminislack-\baselineskip \fi}
\g@addto@macro\WF@finale{\global\wrapminislack-\maxdimen}
\DeclareCaptionFont{wrapcapfont}{\hyphenpenalty=10000 \exhyphenpenalty=10000 }
\newlength{\wrapminiwidth}
\newsavebox{\wrapminibox}
\newcount\wrapminilines
\newcommand{\wrapmini}[4][0]{%
  \par
  \sbox{\wrapminibox}{\parbox[t]{\wrapminiwidth}{%
    \includegraphics[scale=1.13]{#2}\par
    \captionsetup{font={small,wrapcapfont},skip=3pt}\captionof{figure}{#3}\label{#4}}}%
  \wrapminilines=\numexpr(\dimexpr\ht\wrapminibox+\dp\wrapminibox+4pt\relax+\baselineskip/2-1)
    /\baselineskip\relax
  \global\wrapminislack=\dimexpr\wrapminilines\baselineskip-\ht\wrapminibox-\dp\wrapminibox\relax
  \Needspace{\dimexpr\numexpr#1+\wrapminilines\relax\baselineskip+7pt\relax}%
  \global\WF@toplines=#1\relax
  \begin{wrapfigure}[\numexpr#1+\wrapminilines\relax]{r}{\wrapminiwidth}%
    \vspace*{#1\baselineskip}\usebox{\wrapminibox}%
  \end{wrapfigure}}
\newcommand{\wrapminitex}[4][0]{%
  \par
  \sbox{\wrapminibox}{\parbox[t]{\wrapminiwidth}{%
    \input{fig/#2}\par
    \captionsetup{font={small,wrapcapfont},skip=3pt}\captionof{figure}{#3}\label{#4}}}%
  \wrapminilines=\numexpr(\dimexpr\ht\wrapminibox+\dp\wrapminibox+4pt\relax+\baselineskip/2-1)
    /\baselineskip\relax
  \global\wrapminislack=\dimexpr\wrapminilines\baselineskip-\ht\wrapminibox-\dp\wrapminibox\relax
  \Needspace{\dimexpr\numexpr#1+\wrapminilines\relax\baselineskip+7pt\relax}%
  \global\WF@toplines=#1\relax
  \begin{wrapfigure}[\numexpr#1+\wrapminilines\relax]{r}{\wrapminiwidth}%
    \vspace*{#1\baselineskip}\usebox{\wrapminibox}%
  \end{wrapfigure}}
\makeatother
\usepackage{float}
\usepackage{xcolor}
\usepackage{multirow}
\usepackage{subcaption}
\usepackage{tikz}
\usetikzlibrary{calc,positioning,shapes,arrows.meta,fit,backgrounds}
\usepackage{pgfplots}
\usepgfplotslibrary{groupplots}
\pgfplotsset{compat=1.18}

\graphicspath{{fig/}}

\usepackage{xurl}
\usepackage{booktabs}
\usepackage{array}
\newcolumntype{L}[1]{>{\raggedright\arraybackslash}p{#1}}
\usepackage{amsfonts}
\usepackage{amsmath}
\usepackage{amssymb}
\usepackage{mathtools}
\usepackage{nicefrac}
\usepackage{microtype}
\usepackage{makecell}
\usepackage{adjustbox}
\usepackage{placeins}
\usepackage{aliascnt}
\usepackage[capitalize,noabbrev]{cleveref}

\usepackage{xspace}

\theoremstyle{plain}
\newtheorem{theorem}{Theorem}
\newaliascnt{proposition}{theorem}\newtheorem{proposition}[proposition]{Proposition}\aliascntresetthe{proposition}\crefalias{proposition}{proposition}
\newaliascnt{lemma}{theorem}\newtheorem{lemma}[lemma]{Lemma}\aliascntresetthe{lemma}\crefalias{lemma}{lemma}
\newaliascnt{corollary}{theorem}\newtheorem{corollary}[corollary]{Corollary}\aliascntresetthe{corollary}\crefalias{corollary}{corollary}
\theoremstyle{definition}
\newaliascnt{definition}{theorem}\aliascntresetthe{definition}\crefalias{definition}{definition}
\theoremstyle{remark}
\newaliascnt{remark}{theorem}\aliascntresetthe{remark}\crefalias{remark}{remark}
\crefname{remark}{Remark}{Remarks}\Crefname{remark}{Remark}{Remarks}
\theoremstyle{plain}

\newcommand{\E}{\mathbb{E}}

\newcommand{\1}{\mathbf{1}}

\DeclareMathOperator*{\argmax}{arg\,max}

\tcbuselibrary{breakable}
\newtcolorbox{findingsbox}[1]{%
  breakable,fontupper=\small,colback=abstrabg!38,colframe=abstrabg!38,boxrule=0pt,arc=3pt,
  left=7pt,right=7pt,top=5pt,bottom=5pt,
  fonttitle=\bfseries,coltitle=reportaccent,colbacktitle=abstrabg!38,
  title={#1},toptitle=2pt,bottomtitle=0pt}



\hypersetup{pdftitle={Route Anything: A Study of Model and Policy Routing for General-Purpose AI Systems},
  pdfauthor={Pengfei Zhou, Yixing Ma, Hexin Wang, Weidong Tang, Zhenglin Wan, Wangbo Zhao, Yang You},
  pdfsubject={Route Anything},
  pdfkeywords={omni-modal routing, model routing, image generation, video generation, speech generation, robot policies, quality floor, cascades}}

\setheadertext{Route Anything}

\newcommand{\reportinstitutionlogo}[2]{%
  \raisebox{\dimexpr(24pt-\height)/2\relax}{%
    \includegraphics[width=#1,height=24pt,keepaspectratio]{#2}}}
\renewcommand{\institutionlogos}{%
  \hspace{10pt}%
  \reportinstitutionlogo{52pt}{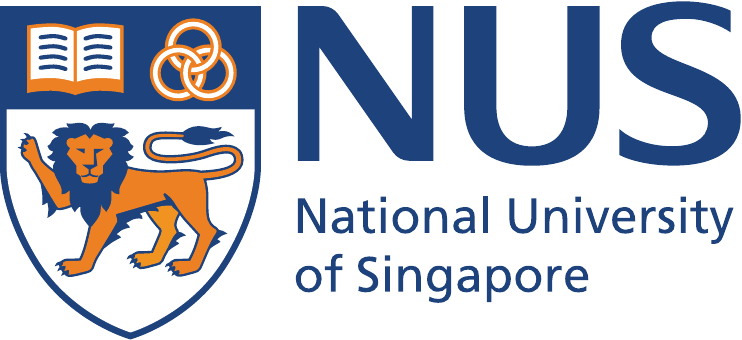}%
  \hspace{10pt}%
  \reportinstitutionlogo{24pt}{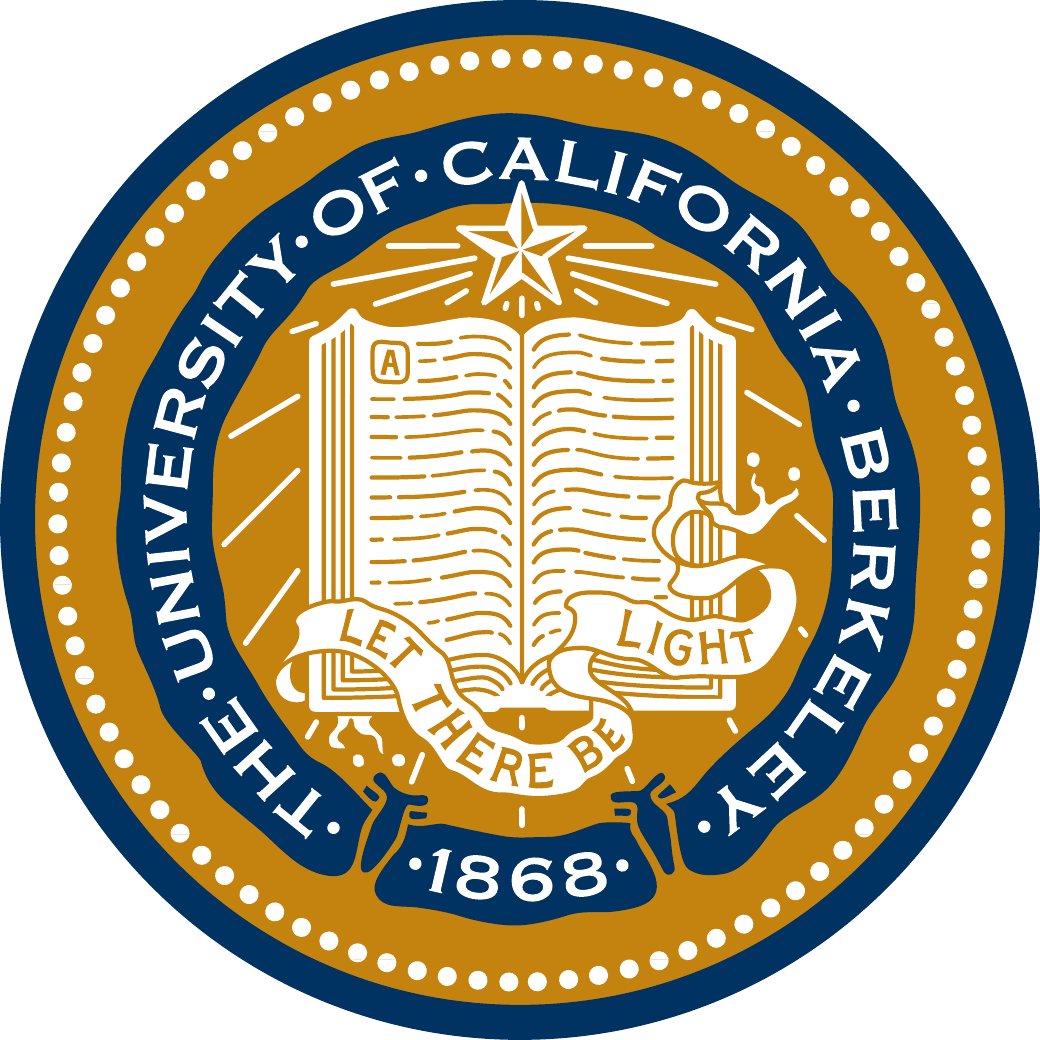}%
  \hspace{4pt}%
  \reportinstitutionlogo{50pt}{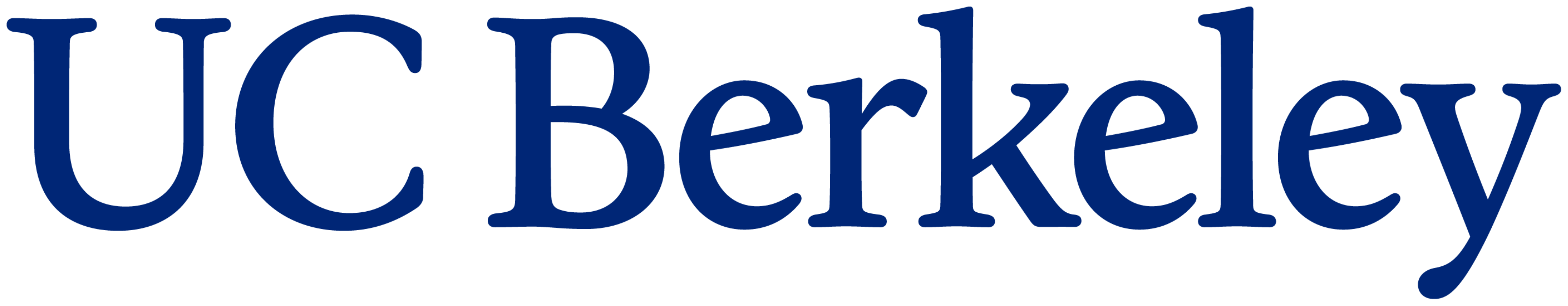}%
  \hspace{10pt}%
  \reportinstitutionlogo{72pt}{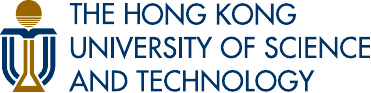}%
}

\title{%
  Route Anything: A Study of Model and Policy Routing for General-Purpose AI Systems
}

\setheadertitle{Route Anything: A Study of Model and Policy Routing for General-Purpose AI Systems}

\author{\parbox{\textwidth}{\centering
    Pengfei Zhou$^{1,2}$ \quad
    Yixing Ma$^{2,3}$ \quad
    Hexin Wang$^{2}$$^\ddagger$ \quad
    Weidong Tang$^{1}$ \quad
    Zhenglin Wan$^{1}$ \quad
    Wangbo Zhao$^{4}$ \quad
    Yang You$^{1}$$^\ddagger$\par\vspace{0.35em}
    {\Affilfont
     \mbox{$^{1}$National University of Singapore, HPC-AI Lab}\hspace{6pt}%
     \mbox{$^{2}$InfRec, Cardinal AI Lab}\hspace{6pt}%
     \mbox{$^{3}$University of California, Berkeley}\hspace{6pt}%
     \mbox{$^{4}$Hong Kong University of Science and Technology}\par}}}

\correspondingemail{\emailicon~\href{mailto:wanghexin@infrec.tech}{wanghexin@infrec.tech}; \emailicon~\href{mailto:yangyou@nus.edu.sg}{yangyou@nus.edu.sg} \quad $^\ddagger$ Corresponding Author}

\newcommand{\projectpageurl}{https://mikeeeeeeeeeeeeee.github.io/RouteAnything/}
\newcommand{\githuburl}{https://github.com/InfRec-Cardinal/route-anything}
\newcommand{\huggingfaceurl}{https://huggingface.co/datasets/Lance1573/route-anything-data}
\makeatletter
\newcommand{\coverlink}[3]{%
  \href{#2}{\tikz[baseline=(chip.base)]\node[draw=black!28,line width=0.5pt,fill=white,
    rounded corners=5.2pt,inner xsep=6.5pt,inner ysep=2.4pt,text=black!85,font=\linkfont\fontsize{8.5}{10}\selectfont]
    (chip){\strut\textcolor{black!80}{#1}\hspace{0.42em}#3};}}
\renewcommand{\lumia@githublink}{%
  \vskip1.5pt
  {\noindent\makebox[\linewidth][l]{\makebox[\dimexpr\linewidth+24pt\relax][c]{%
    \coverlink{\faGlobe}{\projectpageurl}{Project Page}\hspace{6pt}%
    \coverlink{\faGithub}{\githuburl}{GitHub}\hspace{6pt}%
    \coverlink{\faHuggingface}{\huggingfaceurl}{Hugging Face}}}\par}}
\renewcommand{\lumia@huggingfacelink}{}
\makeatother

\let\coverabstract\abscontent
\renewcommand{\abscontent}{%
  {\centering\small
   \textit{AGI should be all-knowing and all-capable. \ \ ---\ Anonymous}\par}%
  \coverabstract}

\renewcommand{\absfont}{\linespread{1.05}\fontsize{10}{12}\selectfont}

\begin{document}

\begin{abstract}
Artificial general intelligence (AGI) requires a method that can understand, generate, reason, and act across diverse domains and modalities. 
Rather than relying on a single model to master every capability, a promising path toward AGI now is to orchestrate a heterogeneous ecosystem of specialized models, tools, and policies (candidates) into a unified AI system. 
At the heart of this vision lies a fundamental question: how can an AI system dynamically select and coordinate the most capable candidate for each task while balancing performance and computational cost?
In this work, we present Route Anything, a unified framework for studying model and policy routing across general-purpose AI systems. 
We develop a unified theoretical foundation that characterizes optimal routing under quality–cost trade-offs and establishes how the information available to a router determines its attainable performance. 
To systematically investigate routing across diverse capabilities, we first construct an omni-modal trajectory matrix spanning 25 benchmarks across language understanding, image, video, and speech generation, coding and workplace agents, and robotic manipulation, comprising 374,401 scored runs from 452 candidate configurations. 
Building on this framework, we train and test different routers to investigate the potential of omni-modal model routing, the capabilities of learned routers, and the conditions under which routing improves efficiency.
Our results reveal substantial opportunities for capability orchestration: per-input oracle routing outperforms the best fixed candidate by 5.4 points, and learned routers outperform it by 0.3 points across 16 benchmarks, by 2.1 points on video generation, and by 1.1 points on robot manipulation while using 24\% less policy time. More importantly, routing raises efficiency by $8.82\times$ on average compared with the reference model, including $25.5\times$ on coding tasks and $7.1\times$ on text question answering.

\end{abstract}

\vspace{3em}

\teaserfigure{%
\vskip4pt
\centering
\includegraphics[width=\linewidth]{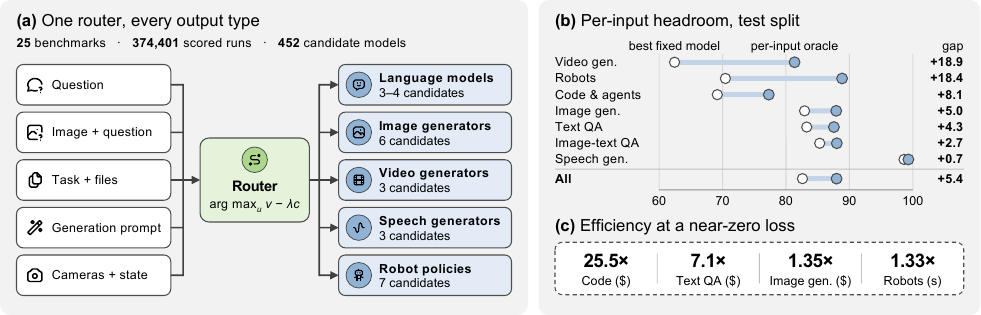}
{\captionsetup{font=small,skip=6pt}
\captionof{figure}{\textbf{Routing across modalities.} (a)~For each input, the router picks one candidate from the pool based on demand, ranking candidates by value minus weighted cost. (b)~On the test set, the best candidate per input (oracle) beats the best fixed candidate per benchmark by 5.4 points on average (Table~\ref{tab:headroom}). (c)~Routing raises efficiency by $8.82\times$ on average over the reference model, at a quality loss below 0.5 points. The efficiency gain, \(\eta/\eta_B\), compares quality per dollar (or per second of policy inference for robots) against the reference (Table~\ref{tab:price}).}
\label{fig:teaser}}
\vskip4pt}

\maketitle

\clearpage
\begingroup
\hypersetup{linkcolor=reporttoc,linktoc=section}
\setlength{\parskip}{0pt}
\setcounter{tocdepth}{2}
\makeatletter\patchcmd{\l@section}{1.0em}{0.5em}{}{}\makeatother
\tableofcontents
\endgroup
\clearpage

\section{Introduction}
\label{sec:intro}

Agentic AI systems now can draw on multiple kinds of specialized models, tools, and policies: a language model answers a question, an image or video generator renders a prompt, a speech model reads text aloud, and a learned policy controls a robot. Rather than relying on one model to master every capability, a general-purpose system can orchestrate this heterogeneous ecosystem and choose, for each input, which module handles it. We call each such model, tool, or policy a \emph{candidate}. The best candidate changes from input to input, so a router that chooses well is expected to match one fixed strongest candidate at a fraction of its price, or beat it by sending every input to the most suitable candidate that handles it best.

Routing research has mostly studied language models: cascades that escalate to a stronger model~\citep{chen2023frugalgpt}, routers learned from preferences~\citep{ong2024routellm}, and quality--cost formulations that connect the two~\citep{dekoninck2025unified}. Whether routing helps across modalities remains an open question. Thus we present a unified framework \emph{Route Anything} for studying model and policy routing across modalities. To support such anlyasitical framework, we first build an omni-modal trajectory matrix in which every candidate in a pool runs on the same inputs, for 25 benchmarks across text and image understanding, image, video, and speech generation, code and workplace agents, and robot manipulation, with three to seven candidates per pool and 374,401 scored runs of 452 candidate configurations (Figure~\ref{fig:teaser}). We train and test routers using different strategies and ask three questions: \emph{how much can routing gain, how well do learned routers work, and when is a cheaper route good enough?}

Our unified routing theory frames all three: a router's value is the best expected outcome, net of priced cost, given what it observes, so what it can reach is a ladder from the best fixed candidate to the per-input oracle, and a reference policy with a quality tolerance fixes the price of quality. The matrix measures each rung, and two code and text testbeds with recorded prices~\citep{zhou2026agent,hu2024routerbench} add the cost side. We also provide empirical insights through extensive experiments:

\begin{findingsbox}{Key findings}
\begin{enumerate}[label=\textbf{\arabic*.},leftmargin=1.5em,itemsep=2pt,topsep=2pt]
\item \textbf{The potential gain for router is large} (\S\ref{sec:headroom}). On 19,814 test inputs from 24 benchmarks, the per-input oracle beats the best fixed candidate per benchmark by 5.4 points, from 0.7 for speech generation to 18--19 for video generation and robot manipulation. The best candidate per capability type, even chosen in hindsight, closes only 8\% of the gap.
\item \textbf{Learned routers beat the best fixed candidate by 0.3 points, and by up to 2.1 points per output group} (\S\ref{sec:one-router}). One capability-tree router over 16 benchmarks scores 79.98 against 79.68 for the fixed candidate and 84.80 for the oracle. Its gains come from agents, video, and image generation; none of 124 router configurations on GenEval2 and Qwen-Image-Bench beats the fixed generator significantly.
\item \textbf{On robots, the router can also read the scene and saves time} (\S\ref{sec:robot}). From three camera frames it sends clean scenes to PatchWAM and randomized ones to ME-Dex, matching the better policy in each condition: 71.6\% success against 70.5\% for the best single policy, with 24\% less policy time.
\item \textbf{Learn outcomes, not winners} (\S\ref{sec:objectives}). With the same inputs and backbone, regressing each policy's success reaches 71.6\% robot success and classifying the winner 42.8\%, because a latency tie-break decides 94\% of the winner labels. Winner-label routers on 16 benchmarks and a 68-type classifier also stay below the fixed candidate.
\item \textbf{Compare candidates on the same inputs} (\S\ref{sec:matched}). A fixed baseline that required a score on every training input dropped Opus 5.5, which missed 11 of 1,122 HLE scores, and turned a learned router's 0.4-point deficit into a 2.4-point gain.
\item \textbf{A cheaper route is often good enough} (\S\ref{sec:policy-choice}). Against the reference, routing raises efficiency by $8.82\times$ on average over four testbeds, including $25.5\times$ on code and $7.1\times$ on text QA. On CodeRouterBench the router fitted for zero quality loss never calls the Opus reference; the certified zero-loss floor holds on test in all 200 splits at a median $19.4\times$, the fitted one in 12.5\%. In image generation, a price route cuts cost by 26\% within half a point.
\item \textbf{Richer candidates and finer states are worth their cost only if they beat what they replace} (\S\ref{sec:spend-more}). On CodeRouterBench, a failure roughly halves the next model's success; a free compiler gate gives $4.09\times$ dollar efficiency, a list-price LLM judge $0.30\times$.
\end{enumerate}
\end{findingsbox}

\paragraph{Contributions.}
\textbf{(1) A unified theory of routing.}~We connect resource prices and quality constraints to optimal model and policy selection, and characterize how the information available to a router determines its attainable performance (\S\ref{sec:model}).
\textbf{(2) An omni-modal trajectory matrix.}~We run 452 candidate configurations on the same inputs of 25 benchmarks across text and image understanding, image, video, and speech generation, code and workplace agents, and robot manipulation, for 374,401 scored runs (\S\ref{sec:headroom} and Appendix~\ref{app:matrix}).
\textbf{(3) The potential of omni-modal routing.}~We quantify the potential gain of routing (5.4 points over the best fixed candidate), train and test learned routers that beat the best fixed candidate by 0.3 points overall and 2.1 points on video generation, show that task-type labels capture 8\% of the potential gain on benchmarks with such labels, and examine how training targets and baseline choices affect measured gains (\S\ref{sec:headroom} and \S\ref{sec:routers}).
\textbf{(4) When routing improves efficiency.}~We show that routing raises efficiency over the reference by $8.82\times$ on average and up to $25.5\times$ on code with matched quality, and evaluate on diverse conditions (\S\ref{sec:price}).

\section{A Router's Value}
\label{sec:model}

Sections~\ref{sec:headroom} and~\ref{sec:routers} use the information ladder and the training targets (Propositions~\ref{prop:ladder} and~\ref{prop:targets}), mostly at $\lambda=0$ since most of the matrix records no costs; Section~\ref{sec:price} adds prices, a reference, and a certified floor (Lemma~\ref{thm:surplus}, Theorem~\ref{thm:optimal-routing}, and Proposition~\ref{prop:certified-floor}).

\paragraph{Candidates, states, and outcomes.}
An input $x$ (a question, an image with a question, a task with files, a generation prompt, or camera frames with a robot state) comes with a finite menu $\mathcal U(x)$ of \emph{candidates}: language models, image, video, or speech generators, robot policies, or execution plans such as ``try a cheap model, verify, and escalate on rejection''. Running candidate $u$ delivers a value $V_u\in[0,1]$, the benchmark's metric, and consumes $D_u>0$ units of one resource, such as dollars or seconds, with $\E[D_u]<\infty$. A router observes a \emph{state} $s=\sigma(x)$, which may be the benchmark, a capability type, or the input itself, and chooses a candidate $g(s)$, possibly at random. We write $v(s,u)=\E[V_u\mid s]$ and $c(s,u)=\E[D_u\mid s]$. Routing does not change which inputs arrive, so the law $\pi$ of states is fixed; sequential choices within a task enter as complete plans.

\paragraph{Price, reference, and floor.}
At a resource price $\lambda\ge0$, router $g$ earns $J_\lambda(g)=\E[V_g]-\lambda\E[D_g]$; at $\lambda=0$ it is ranked by quality alone. When no price is given, we compare $g$ with a named \emph{reference} policy, such as always calling the strongest model, whose value and cost on the same inputs are $Y_B$ and $C_B$, and define the quality change $Q$, the saving $\mathcal S$, the efficiency $\eta$, and the quality floor with tolerance $\epsilon\ge0$:
\begin{equation}
Q=\E[V_g-Y_B],\quad \mathcal S=\E[C_B-D_g],\quad \eta=\frac{\E[V_g]}{\E[D_g]},\quad \eta_B=\frac{\E[Y_B]}{\E[C_B]},\quad Q\ge-\epsilon .
\label{eq:main-objective}
\end{equation}
\begin{lemma}[Reference-relative gain]
\label{thm:surplus}
$\eta-\eta_B=(Q+\eta_B\mathcal S)/\E[D_g]$. Efficiency improves if and only if $Q+\eta_B\mathcal S>0$, and the quality floor is the separate condition $Q\ge-\epsilon$.
\end{lemma}
Both conditions are needed: a primary that is correct on 15\% of tasks at a tenth of the cost of an always-correct reference has $\eta=1.5\,\eta_B$ but $Q=-0.85$.

\begin{theorem}[Optimal routing at a price and under a floor]
\label{thm:optimal-routing}
Let the state take finitely many values, each with $\pi(s)>0$, let the menu depend on $x$ only through $s$, and let costs be positive.
(i)~A router maximizes $J_\lambda$ if and only if, in every state, it plays only candidates maximizing $v(s,u)-\lambda c(s,u)$. The highest efficiency $\eta^\star=\max_g\eta(g)$ is the unique root of $\Phi(\lambda)=\sum_s\pi(s)\max_u[v(s,u)-\lambda c(s,u)]$, and the efficiency-optimal routers are exactly the maximizers of $J_{\eta^\star}$.
(ii)~If the reference is itself a router and $\epsilon\ge0$, an efficiency-optimal router under the floor $Q\ge-\epsilon$ exists, and for some $\kappa\ge0$ every such router maximizes $J_\lambda$ at $\lambda_\epsilon=\eta^\star_\epsilon/(1+\kappa)$, where $\eta^\star_\epsilon$ is the constrained optimum.
\end{theorem}
The proof combines Dinkelbach's method~\citep{dinkelbach1967nonlinear} with linear-programming duality (Appendix~\ref{app:proof-optimal-routing}). Part~(ii) says that maximizing efficiency under a floor yields no new kind of router: it is routing at one price, set by the reference and the tolerance rather than by hand or by a budget~\citep{dekoninck2025unified}. Without recorded costs, the rule is to play the candidate with the highest expected value given the state.

\paragraph{The information ladder.}
What a router can reach depends on what it observes. Write $W_u=V_u-\lambda D_u$ and let $V^\star(\sigma)=\E\bigl[\max_u\E[W_u\mid\sigma(x)]\bigr]$ be the best return of a router that observes $\sigma(x)$, attained, as in part~(i) of Theorem~\ref{thm:optimal-routing}, by playing in each state a candidate that maximizes $\E[W_u\mid\sigma(x)]$.
\begin{proposition}[Information ladder]
\label{prop:ladder}
Let $\sigma_1$ be coarser than $\sigma_2$, that is, $\sigma_1(x)$ is a function of $\sigma_2(x)$, and let the menu depend on $x$ only through $\sigma_1(x)$. Then
\begin{equation}
V^\star(\sigma_1)\ \le\ V^\star(\sigma_2)\ \le\ V^\star(x)\ \le\ \E\bigl[\max_uW_u\bigr].
\label{eq:ladder}
\end{equation}
The first inequality is an equality if and only if, for almost every value $s_1$ of $\sigma_1$, a single candidate is optimal at almost every finer state $s_2$ with $\sigma_1=s_1$. The last gap equals $\E\{\E[\max_uW_u\mid x]-\max_u\E[W_u\mid x]\}$, and it vanishes if and only if, given almost every $x$, one candidate attains $\max_uW_u$ almost surely.
\end{proposition}
With $\sigma_1$ the benchmark, $V^\star(\sigma_1)$ is the best fixed candidate per benchmark. The outer rungs are measurable: the best fixed candidate on the left and the per-input oracle on the right. Inner rungs are bounded below by any router that observes that state and estimated optimistically by the best lookup chosen on the test inputs. A coarse state helps only where it changes the best candidate, and the last gap is outcome noise that no observable state predicts, as when a candidate's success on an input is a coin flip.

\paragraph{What to learn.}
A learned router needs a training target. Two are common: each candidate's outcome, and the identity of the winning candidate.
\begin{proposition}[Outcomes, not winners]
\label{prop:targets}
(a)~For each candidate, the score that minimizes the expected binary cross-entropy, or squared error, against $V_u$ given $x$ is $m_u(x)=\E[V_u\mid x]$, and routing to $\argmax_um_u(x)$ attains $V^\star(x)$ at $\lambda=0$. (b)~Let the \emph{winner} be the candidate with the highest realized value, ties broken by a fixed rule, and let $p_u(x)$ be the probability that $u$ wins, which a cross-entropy classifier on winner labels estimates. Routing to $\argmax_up_u(x)$ can lose almost half of the value range: if candidate $A$ always delivers $a<1$ and candidate $B$ delivers $1$ with probability $b\in(\tfrac12,a)$ and $0$ otherwise, then $p_B=b>p_A$, while $\E[V_B]=b<a$.
\end{proposition}
Tie-breaks make it worse: with binary outcomes and ties sent to a faster candidate, a candidate that succeeds with probability 0.9 loses the label to a faster independent candidate that succeeds with probability 0.5 (it wins with probability 0.45). Per-candidate regression uses every outcome, including ties and failures, whereas a winner label keeps one choice per input.

\paragraph{Certifying the floor.}
Theorem~\ref{thm:optimal-routing} is applied to fitted values and costs, so a fitted floor need not hold on test inputs.
\begin{proposition}[Certified quality floor]
\label{prop:certified-floor}
Let values lie in $[0,1]$, let $\epsilon\ge0$ and $\alpha\in(0,1)$, and let the routers $g_1,\dots,g_K$ and their order be fixed by the fitting data. Let $p_k$ be valid $p$-values for the null hypotheses $Q_{g_k}<-\epsilon$, computed on $n$ inputs drawn i.i.d.\ from the deployment law, independently of the fitting data, with matched outcomes; for instance $p_k=\exp\{-n(\hat Q_k+\epsilon)_+^2/2\}$, where $\hat Q_k$ is the sample mean of $V_{g_k}-Y_B$. Let $\hat g=g_{\hat k}$ with $\hat k=\max\{k:p_1,\dots,p_k\le\alpha\}$, or the reference if $p_1>\alpha$. Then $\mathbb P(Q_{\hat g}<-\epsilon)\le\alpha$.
\end{proposition}
\wrapmini[0]{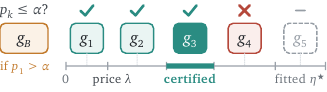}{\textbf{Certifying a floor.} Routers on the price path are tested in order on $\mathcal B$.}{fig:mini-path}
A natural order is the \emph{price path} (Figure~\ref{fig:mini-path}): the maximizers of the fitted $J_\lambda$ as $\lambda$ rises from 0 to the fitted $\eta^\star$, along which fitted value and cost fall while fitted efficiency rises (Appendix~\ref{app:proof-certified-floor}). The test stops at the first router it cannot certify, so no multiplicity correction is needed, as in fixed-sequence testing~\citep{westfall2001optimally,angelopoulos2025learn}; returning the reference when nothing is certified mirrors high-confidence policy improvement~\citep{thomas2015high}. The certified router comes with a certified price: the reference, the tolerance, and the level fix $\lambda$.

\section{Can Router Beat the Best Fixed Candidate?}
\label{sec:headroom}

\begin{table}[t]
\centering\small\setlength{\tabcolsep}{5pt}
\caption{\textbf{Per-input potential gain on the test split.} Inputs on which every candidate of the pool is scored; values are benchmark metrics on a 0--100 scale, pooled over inputs. \emph{Fixed} plays the best candidate per benchmark on training inputs, \emph{best fixed in hindsight} the best candidate per benchmark on these test inputs, and the \emph{oracle} the best candidate per input.}
\label{tab:headroom}
\begin{tabular}{@{}lrrrrrrr@{}}
\toprule
Group & Benchmarks & Candidates & Inputs & Fixed & \makecell{Best fixed\\in hindsight} & Oracle & Gap\\
\midrule
Text QA & 7 & 3--4 & 9,482 & 83.27 & 83.27 & 87.56 & 4.28\\
Image-text QA & 3 & 3 & 5,708 & 85.32 & 85.32 & 88.02 & 2.70\\
Image generation & 5 & 4--6 & 1,515 & 82.93 & 82.99 & 87.92 & 4.99\\
Video generation & 1 & 3 & 158 & 62.43 & 62.43 & 81.36 & 18.93\\
Speech generation & 2 & 3 & 608 & 98.62 & 98.63 & 99.30 & 0.68\\
Code and workplace agents & 5 & 4 & 343 & 69.17 & 69.78 & 77.29 & 8.11\\
Robot manipulation & 1 & 7 & 2,000 & 70.45 & 70.45 & 88.85 & 18.40\\
\midrule
All & 24 & & 19,814 & 82.60 & 82.62 & 87.98 & 5.38\\
\bottomrule
\end{tabular}
\end{table}

\paragraph{Setup.}
Every candidate in a pool runs on the same inputs, and each run is scored by the benchmark's own metric, rescaled to $[0,1]$ and reported on a 0--100 scale. The 25 benchmarks fall into seven groups (Table~\ref{tab:headroom}): text QA~\citep{wang2024mmlupro,gema2025mmluredux,du2025supergpqa,rein2024gpqa,aa2026evals,jackson2025omniscience,phan2025hle}, image-text QA~\citep{yue2025mmmupro,yue2024mmmu,yu2026scope}, image generation~\citep{kamath2025geneval2,li2026qwenimagebench,yang2026longt2ibench,chang2025oneig,niu2025wise}, video generation~\citep{zheng2025vbench2}, speech generation~\citep{anastassiou2024seedtts}, code and workplace agents~\citep{shepard2026automationbench,garre2026gdppdf,patwardhan2025gdpval,tian2024scicode,tbench2025}, and robot manipulation~\citep{chen2025robotwin2}; Appendix~\ref{app:matrix} lists them with their metrics and candidates. The language-model candidates are Claude Opus~5.5, GPT-6.1 Sol, and DeepSeek V4.1 Flash, joined by GPT-6 Astra on long-context, knowledge, and agent tasks. Inputs are split 70/10/20 into training, development, and test sets by source group, so identical inputs and items from one source never cross splits; RoboTwin's 2,000 test scenes come from 10 tasks the routers never see in training. The \emph{fixed candidate} plays the best candidate per benchmark on training inputs, where missing outcomes never disqualify a candidate (Section~\ref{sec:matched} shows why this matters), and the \emph{oracle} plays the best observed candidate on each input. These are the outer rungs of Proposition~\ref{prop:ladder}. We measure them on the 19,814 test inputs on which every candidate is scored, which leaves out SciCode, whose later steps contain each model's own earlier code, and estimate the rungs in between with the best lookup per benchmark or per type chosen in hindsight.

\wrapmini[0]{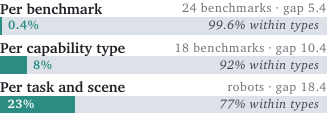}{\textbf{Where the potential gain lies.} Gap recovered by the best candidate per type.}{fig:mini-headroom}
\paragraph{Finding 1: The potential gain is large, and it lies within task types.}
On the 19,814 test inputs, the best fixed candidate per benchmark scores 82.60 and the per-input oracle 87.98 (Table~\ref{tab:headroom}). The gap is 0.7 points for speech generation, where every generator is nearly perfect, and 18--19 points for video generation and robot manipulation, where no candidate dominates. Little of it is available to a router that sees only the benchmark or the task type. The best candidate per benchmark chosen in hindsight raises 82.60 only to 82.62. On the 18 benchmarks of the rule-based capability tree, which assigns each of 5,034 test inputs to one of 107 capability leaves (Section~\ref{sec:one-router}), the best candidate per leaf chosen in hindsight scores 77.89, against 77.03 for the fixed candidate and 87.44 for the oracle: 8\% of the gap lies between types. On RoboTwin, the best policy per task and scene condition, again chosen on the test scenes, recovers 23\% of the 18.4-point gap. Code tasks show the same split. The best released CodeRouterBench routers capture 55\% of the value of perfect routing information when the task type is unknown, but only 6--8\% within a task type (Appendix~\ref{app:signal-details}).

Inside a type, outcomes vary input by input. On GenEval2, each of the five other generators beats the fixed one by more than two points on 76--100 training prompts while trailing it on more, and the per-prompt difference has a standard deviation of 31--38 points against mean differences of 4--14. Part of this variation is noise that no router can predict (the last gap of Proposition~\ref{prop:ladder}). On AA-LCR, where every question is answered three times, the oracle's 2.4-point margin over the fixed candidate vanishes once each candidate is scored by its mean over the three runs.

Section~\ref{sec:routers} studies learned routers on every benchmark except seven understanding ones (MMLU-Pro, MMLU-Redux, SuperGPQA, GPQA, and the three image-text QA benchmarks), which hold three quarters of the inputs and half of the pooled gap in Table~\ref{tab:headroom}, with gaps of 1.4--6.0 points; every benchmark with a gap above 8 points is studied.

\section{How Well Do Learned Routers Work?}
\label{sec:routers}

Our routers fine-tune Qwen3.5-4B on the task text and the output requirement, and for robots also on three camera frames and the robot state (Appendix~\ref{app:routers}). An \emph{outcome router} scores each pair of input and candidate profile with a sigmoid head, trained by cross-entropy against the candidate's normalized outcome (Proposition~\ref{prop:targets}(a)), and plays the highest score. A \emph{type router} classifies the input into a capability type and plays that type's best candidate, read from a table fitted on training inputs, or the fixed candidate when it cannot place the input. A \emph{winner-label router} is trained to name the best candidate. Each study is scored on the test inputs where all its compared policies have outcomes (Table~\ref{tab:panels}); differences carry paired 95\% bootstrap intervals over source groups, or over scenes for RoboTwin.

\begin{table}[t]
\centering\footnotesize\setlength{\tabcolsep}{4pt}
\caption{\textbf{Evaluation panels.} Each study uses the test inputs on which all its compared policies are scored; fixed-candidate and oracle scores (0--100) are comparable only within a row. The image-routing suite of Section~\ref{sec:one-router} uses 158 GenEval2 and 193 Qwen-Image-Bench prompts, and the image price routes of Section~\ref{sec:policy-choice} use the 1,515 image-generation inputs of Table~\ref{tab:headroom}.}
\label{tab:panels}
\begin{tabular}{@{}llrrrrr@{}}
\toprule
Study & Policies compared & Benchmarks & Inputs & Seeds & Fixed & Oracle\\
\midrule
Potential gain (\S\ref{sec:headroom}) & fixed candidate, hindsight lookups & 24 & 19,814 & -- & 82.60 & 87.98\\
Rule-based tree (\S\ref{sec:headroom}, \S\ref{sec:one-router}) & rule-based type router & 18 & 5,034 & -- & 77.03 & 87.44\\
One router (\S\ref{sec:one-router}) & capability-tree and untrained routers & 16 & 3,444 & 1 & 79.68 & 84.80\\
Robots (\S\ref{sec:robot}, \S\ref{sec:objectives}) & outcome and winner-label routers & 1 & 2,000 & 1 & 70.45 & 88.85\\
Targets (\S\ref{sec:objectives}, \S\ref{sec:matched}) & outcome and type routers & 11 & 1,302 & 4 & 68.30 & 75.84\\
Winner labels (\S\ref{sec:objectives}) & 12 winner-label recipes & 16 & 3,211 & 1 & 79.44 & 85.77\\
\bottomrule
\end{tabular}
\end{table}

\subsection{One Router Across Modalities}
\label{sec:one-router}

\paragraph{Setup.}
The capability-tree router~\cite{tang2026trinity} is a type router whose types are the 157 leaves of a tree over input modality, task, and capability, on 18 benchmarks: all except MMLU-Pro, MMLU-Redux, SuperGPQA, GPQA, and the three image-text QA benchmarks. A Qwen3.5-4B classifier, trained on all benchmarks at once, predicts the leaf, and one global table, fitted on training inputs, maps each leaf to its best candidate; unresolved inputs go to the fixed candidate. We evaluate it on the 3,444 test inputs of 16 benchmarks on which every compared policy is scored (Table~\ref{tab:routers}); the robot benchmark is handled by the outcome router of Section~\ref{sec:robot}.

\begin{table}[t]
\centering\small\setlength{\tabcolsep}{5pt}
\caption{\textbf{Learned routers against the fixed candidate and the oracle.} Test inputs; one training seed. \emph{Share} is the router's gain over the fixed candidate divided by the oracle's. The first block is the capability-tree router; the robot row is the outcome router of Section~\ref{sec:robot}.}
\label{tab:routers}
\begin{tabular}{@{}lrrrrrr@{}}
\toprule
Group (benchmarks) & Inputs & Fixed & Router & Oracle & Router $-$ fixed & Share\\
\midrule
Text QA (3) & 588 & 65.14 & 64.97 & 71.43 & $-0.17$ & $-3\%$\\
Code and workplace agents (5) & 519 & 69.24 & 70.01 & 74.78 & $+0.77$ & 14\%\\
Image generation (5) & 1,559 & 82.96 & 83.21 & 87.92 & $+0.25$ & 5\%\\
Video generation (1) & 161 & 62.17 & 64.28 & 80.22 & $+2.12$ & 12\%\\
Speech generation (2) & 617 & 98.63 & 98.63 & 99.30 & $0.00$ & 0\%\\
\midrule
All 16 benchmarks & 3,444 & 79.68 & 79.98 & 84.80 & $+0.30$ & 6\%\\
Robot manipulation (outcome router) & 2,000 & 70.45 & 71.55 & 88.85 & $+1.10$ & 6\%\\
\bottomrule
\end{tabular}
\end{table}

\paragraph{Finding 2: learned routers beat the best fixed candidate by 0.3 points, and by up to 2.1 points per output group.}
One router over 16 benchmarks in five output groups scores 79.98, against 79.68 for the fixed candidate and 84.80 for the oracle. It matches the fixed candidate to within 0.01 points on 10 benchmarks and departs from it where candidates complement each other input by input: Terminal-Bench ($+8.3$ on 36 inputs), VBench-2.0 ($+2.1$), OneIG ($+1.1$), and GDP.pdf ($+1.0$); its largest loss is AA-LCR ($-2.4$). The leaf classifier is 86--89\% accurate, yet 83--86\% of inputs still go to the fixed candidate, because within a benchmark one candidate wins almost every type (Proposition~\ref{prop:ladder}). A rule-based tree with 107 leaves, which needs no classifier, scores 0.09 points above the fixed candidate, almost all of it on OneIG. In image generation, across 31 router designs and 124 configurations on GenEval2 and Qwen-Image-Bench, none beats the fixed generator with an interval above zero, and 44 are significantly worse. An untrained Qwen3.5-4B that picks the candidate itself scores 1.15 points below the fixed candidate.

\subsection{Routers That Read the Scene}
\label{sec:robot}

\wrapmini[0]{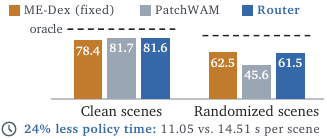}{\textbf{Routing by scene condition.} Robot success on 2,000 test scenes.}{fig:robot}
\paragraph{Setup.}
On RoboTwin, the outcome router reads the instruction, three $240\times320$ camera frames, and the robot state, and predicts the success of each of seven vision-language-action policies with one sigmoid head per policy, trained by cross-entropy on all 49,000 training outcomes. No task identity enters the input, and the 2,000 test scenes come from 10 tasks the router never sees, half in clean scenes and half in domain-randomized ones. ME-Dex is the best policy on training scenes and also on the test scenes (70.45\%); PatchWAM is second (63.65\%). A policy's time on a scene is its inference seconds, summed over its queries.

\paragraph{Finding 3: on robots, the router reads the scene and saves time.}
The router sends 929 of the 1,000 clean scenes to PatchWAM and 842 of the 1,000 randomized scenes to ME-Dex (141 more to GigaBrain). The best policy flips between the two conditions, and the router tracks it: on clean scenes it scores 81.6\%, against 81.7\% for PatchWAM and 78.4\% for ME-Dex; on randomized scenes 61.5\%, against 62.5\% for ME-Dex and 45.6\% for PatchWAM (Figure~\ref{fig:robot}). Overall success is 71.55\% against 70.45\% ($+1.10$ $[-0.40,+2.55]$), at 11.05 instead of 14.51 seconds of policy time per scene, 24\% less: $1.33\times$ the success per policy-second of the best single policy. This is the coarse rung of Proposition~\ref{prop:ladder} learned from pixels, since the condition-level choice alone would score 72.1\%. The remaining 17\% to the oracle require predicting which policy succeeds on a given scene.

\subsection{What a Router Should Learn}
\label{sec:objectives}

\wrapmini[0]{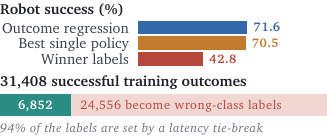}{\textbf{Learn outcomes, not winners.} Winner labels mislabel most successes.}{fig:mini-targets}
\paragraph{Finding 4: learn outcomes, not winners.}
With the same inputs, backbone, and training scenes, the winner-label classifier reaches 42.8\% robot success against 71.6\% for per-policy outcome regression (Table~\ref{tab:objectives}). The labels explain the gap, as Proposition~\ref{prop:targets}(b) predicts: several policies succeed on most training scenes, a latency tie-break decides 94\% of the labels, and 24,556 of the 31,408 successful training outcomes become wrong-class targets, so the classifier learns to pick fast policies (1.9 seconds per scene) that often fail. Outside robotics, no panel compares the two targets directly, and no target beats the fixed candidate. Generative winner-label routers on 16 benchmarks learn from 11,273 labels of which only a third are decided by a unique best outcome, the rest by cost, tokens, or a fixed order; the best of 12 recipes scores 78.87 against 79.44 for the fixed candidate. On 11 text, agent, and image benchmarks, the outcome router stays within noise of the matched fixed candidate ($-0.40$), and a type router over 68 types, although 75\% accurate, maps 93\% of its predictions to the same candidate as the true type and scores 1.32 points below it.

\begin{table}[t]
\centering\small\setlength{\tabcolsep}{4pt}
\caption{\textbf{Training targets and the fixed baseline.} Paired 95\% intervals over source groups (scenes for robots). The first block averages four training seeds on 1,302 test inputs of 11 text, agent, and image benchmarks; the fixed candidate is matched (Section~\ref{sec:matched}).}
\label{tab:objectives}
\begin{tabular}{@{}llrrr@{}}
\toprule
Study & Policy & Inputs & Score & $-$ matched fixed [95\% CI]\\
\midrule
11 benchmarks, 4 seeds & Matched fixed candidate & 1,302 & 68.30 & --\\
 & Outcome router & 1,302 & 67.90 & $-0.40$ $[-1.19,+0.33]$\\
 & 68-type classifier + table & 1,302 & 66.98 & $-1.32$ $[-3.05,+0.34]$\\
 & Oracle & 1,302 & 75.84 & $+7.55$\\
\midrule
16 benchmarks & Fixed candidate (best training mean) & 3,211 & 79.44 & --\\
 & Winner-label router (generative) & 3,211 & 78.87 & $-0.58$\\
 & Oracle & 3,211 & 85.77 & $+6.33$\\
\midrule
Robot manipulation & Outcome router (per-policy success) & 2,000 & 71.55 & $+1.10$ $[-0.40,+2.55]$\\
 & Winner-label classifier & 2,000 & 42.80 & $-27.65$\\
\bottomrule
\end{tabular}
\end{table}

\Needspace{10\baselineskip}
\subsection{Matched Comparisons}
\label{sec:matched}

\wrapmini[0]{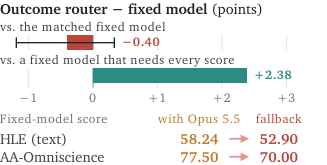}{\textbf{Compare candidates on the same inputs.} Missing scores flip the sign.}{fig:mini-matched}
\paragraph{Finding 5: compare candidates on the same inputs.}
A fixed baseline that admits a model only if it has a score on every training input drops strong models over a few missing scores. In the 11-benchmark study, Opus~5.5 missed a few scores on seven benchmarks, for example 11 of 1,122 training inputs of HLE, so the baseline fell back to the next model there: on HLE it scored 52.90 instead of 58.24, and on AA-Omniscience 70.00 instead of 77.50. Pooled, the baseline dropped from 68.30 to 65.52, and the outcome router's 0.40-point deficit became a 2.38-point gain (Table~\ref{tab:objectives}). Choosing and evaluating the fixed candidate under the routers' fallback, which uses GPT-6 Astra's outcome where Opus~5.5 has none, removes the artifact. The same rule governs portfolios. Adding a model changes which tasks reach the fallback, so averages frozen at the current portfolio misrank candidates, whereas comparing both portfolios on the same tasks picks the best added model in 96.4--97.1\% of CodeRouterBench comparisons, against 86.3--94.3\% for frozen averages (Appendix~\ref{app:portfolio-details}).

\Needspace{8\baselineskip}
\section{When Is a Cheaper Route Good Enough?}
\label{sec:price}

\subsection{The Price of a Route}
\label{sec:policy-choice}

\paragraph{Setup.}
A route trades quality for cost where candidates differ more in price than in value. The matrix records prices for image and speech generation and time for robot policies. For code and text, two testbeds add token prices: CodeRouterBench~\citep{zhou2026agent}, with eight models on 9,999 coding tasks (7,080 probing tasks for fitting and 2,919 test tasks) and the reference claude-opus-4-6 (Opus~4.6), and RouterBench~\citep{hu2024routerbench}, with eleven models on 36,497 mostly non-code prompts split 70/30 and the reference gpt-4-1106-preview. On CodeRouterBench, states are the 21 dimension$\times$source cells, and each state offers 22 candidates: direct dispatch to one of eight models, or one of seven primaries followed by Opus~4.6 when the primary's released score, an oracle verifier, is below $\theta\in\{1,0.5\}$. Such a reference-fallback cascade with verifier decision $H\in\{0,1\}$ and primary-plus-verification cost $A$ delivers
\begin{equation}
V=HY_P+(1-H)Y_B,\qquad D=A+(1-H)C_B .
\label{eq:accounting-main}
\end{equation}
Values, costs, and state frequencies are fitted on probing tasks, the floor $Q\ge-\epsilon$ is imposed there, and policies are evaluated on test tasks; always-Opus scores 0.4383 at 11.655\,m\$ per task. RouterBench uses source-family states and the same construction. For image generation, states are the five benchmarks, candidates are the generators (three on GenEval2, six elsewhere), and the reference plays the best generator per benchmark (the fixed candidate).

\paragraph{When the reference is worth calling.}
A candidate that calls the reference is worth its price only while the reference earns enough on the inputs it receives.
\begin{corollary}[Admission price]
\label{cor:admission}
In the setting of Theorem~\ref{thm:optimal-routing}, let $\lambda^B$ be the largest price at which some candidate that calls the reference maximizes $v(s,\cdot)-\lambda c(s,\cdot)$ in some state. If $\lambda^B<\eta^\star$, the efficiency-optimal router never calls the reference, and neither does a floor-optimal router whose price $\lambda_\epsilon$ exceeds $\lambda^B$. A cascade~\eqref{eq:accounting-main} whose primary is also available by direct dispatch maximizes only at prices up to $\E[(Y_B-Y_P)\1_{H=0}\mid s]/\E[C_B\1_{H=0}\mid s]$, the reference's added value per dollar on the inputs it receives.
\end{corollary}

\begin{table}[t]
\centering\small\setlength{\tabcolsep}{4pt}
\caption{\textbf{Price routes across testbeds.} Test-set quality change $\Delta V=\E[V-Y_B]$ in points (0--100) with paired 95\% intervals, and efficiency $\eta/\eta_B$ against the reference. Fitted routes meet the floor on fitting data; certified routes are medians over 200 random halvings of the fitting data (Proposition~\ref{prop:certified-floor}).}
\label{tab:price}
\begin{tabular}{@{}llllr@{}}
\toprule
Testbed (resource) & Reference & Route & $\Delta V$ [95\% CI] & $\eta/\eta_B$\\
\midrule
Code, CodeRouterBench (\$) & Always Opus~4.6 & Fitted, $\epsilon=0$ & $-0.44$ $[-1.75,+0.97]$ & $25.5\times$\\
 & & Certified, $\epsilon=0$ & $\ge0$ in 200 of 200 & $19.4\times$\\
Text QA, RouterBench (\$) & Always GPT-4 & Fitted, $\epsilon=0$ & $-0.40$ $[-0.84,+0.08]$ & $7.1\times$\\
 & & Certified, $\epsilon=0$ & $\ge0$ in 200 of 200 & $6.7\times$\\
Image generation (\$) & Fixed generator & Fitted, $\epsilon=0.5$ & $-0.47$ $[-0.87,-0.06]$ & $1.35\times$\\
 & & Fitted, $\epsilon=2$ & $-1.56$ $[-2.39,-0.73]$ & $2.03\times$\\
Robot manipulation (s) & Fixed policy & Outcome router & $+1.10$ $[-0.40,+2.55]$ & $1.33\times$\\
\bottomrule
\end{tabular}
\end{table}

\wrapmini[0]{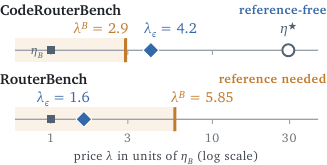}{\textbf{Admission vs.\ floor price} in units of $\eta_B$; the shaded range is $\lambda\le\lambda^B$.}{fig:mini-ruler}
\paragraph{Finding 6: a cheaper route is often good enough.}
Across the four testbeds of Table~\ref{tab:price}, routing raises efficiency over the reference by $8.82\times$ on average (the mean of $25.5$, $7.1$, $1.35$, and $1.33\times$, with image generation at $\epsilon=0.5$). On CodeRouterBench, the admission price of Opus~4.6 is $\lambda^B=100$ score per dollar ($2.9\,\eta_B$), below the zero-loss floor price of 144 and far below $\eta^\star=1022$, so by Corollary~\ref{cor:admission} the zero-loss router never calls Opus. It dispatches every task to a cheaper model and reaches $25.5\times$ Opus's efficiency at a 0.44-point loss (Table~\ref{tab:price}). Always calling Qwen3-Max, the cheapest model, already reaches $20.1\times$ at a 4.2-point loss; routing by state recovers that loss while adding 27\% efficiency. The states need not be oracle labels: prompt-inferred cells reach $25.3\times$, and learned quality--cost routers $22.7$--$24.2\times$ at their zero-loss points (Appendix~\ref{app:policy-details}). Without states, every zero-loss policy calls Opus, which then runs on 28.7\% of tasks. On RouterBench, the best router that avoids the reference scores 9.7 points below it, so the zero-loss router cascades cheap primaries to the reference and reaches $7.1\times$. A fitted floor is not a guarantee. Refitted on half of the probing tasks, the zero-loss floor holds on test in only 12.5\% of 200 splits; certified on the other half by Proposition~\ref{prop:certified-floor} with empirical-Bernstein $p$-values on the mixture-refined path, it holds in all 200, at a median $19.4\times$. Image generation follows the same logic. GPT Image~2.5 costs \$0.21 per image, 3.5--5.4$\times$ the other generators, and is the best candidate on two of the five benchmarks. The route that stays within half a point on training prompts moves Qwen-Image-Bench to Grok Imagine~2.0 and OneIG to Qwen-Image~3, cutting test cost by 26\% at a 0.47-point loss, and a two-point tolerance halves cost. In speech generation the best generator is also among the cheapest, so there is nothing to trade, and on robots the time saving of Section~\ref{sec:robot} comes with no loss of success. Router overhead is small next to these prices: a 4B router takes 0.13s per decision on one H20 GPU, whereas generating an image takes 27--158s and a video 211--314s.

\subsection{Spending More Within a Task}
\label{sec:spend-more}

\paragraph{Setup.}
A candidate can spend more inside a task: a plan that retries after a failure, or that verifies an output and escalates on rejection. A router can also buy a finer state: a forecast of a candidate's success, or a record of past outcomes. By Theorem~\ref{thm:optimal-routing}, a richer candidate is worth playing only if its value net of its full cost beats the candidates it replaces, and that value must be computed on the inputs that reach each step (Proposition~\ref{prop:attempts}). By Proposition~\ref{prop:ladder}, a finer state can only help an optimal router, but it is worth buying only if the help exceeds its price (Proposition~\ref{prop:voi}). We test both on CodeRouterBench, with fresh model calls for verifier verdicts and forecasts, and semi-synthetic streams for memory (Appendix~\ref{app:experiments}).

\paragraph{Finding 7: richer candidates and finer states are worth their cost only if they beat what they replace.}
\emph{Retries.} A failure leaves harder tasks: in all 42 ordered model pairs, the second model succeeds less often after the first fails (median ratio 0.49 at $\theta=1$). \emph{Verifiers.} For Qwen3-Max at $\theta=0.5$, a free compiler gate gives $4.09\times$ dollar efficiency at a 2.4-point loss, and an LLM judge $2.63\times$ if free but $0.30\times$ at its list price; in seconds, always-Opus beats every verified cascade, because Opus answers faster than the primaries. \emph{Finer states.} A forecast can save at most the attempt it replaces: a DeepSeek-v4-pro forecast costs 2.4 attempts at list price, a prompt-only TF-IDF router 0.06\% of one.
\section{Related Work}
\label{sec:related}

\paragraph{Routing language models.}
Cascades and routers trade quality against cost~\citep{chen2023frugalgpt,ong2024routellm,dekoninck2025unified}, and learned routers estimate each model's quality from the query~\citep{ding2024hybrid,chen2024routerdc,feng2025graphrouter,zhuang2025embedllm,ding2025bestroute}. The acceptance step of a cascade is a reject option~\citep{chow1970optimum}, studied as confidence-based or budgeted deferral~\citep{jitkrittum2023does,gupta2024language,aggarwal2024automix,chen2020frugalml,trapeznikov2013supervised}. \citet{dekoninck2025unified} characterize optimal routing and cascading through a Lagrangian price tuned to a budget; here, the reference and the quality tolerance fix that price, and a finite-sample test certifies the resulting floor. Routing benchmarks record per-model outcomes and costs on shared prompts~\citep{hu2024routerbench,zhou2026agent}.

\paragraph{Routing beyond text.}
SCOPE-Router arranges vision-language models for execution-oriented tasks~\citep{yu2026scope}, and routing ideas also inspire vision-token pruning strategies~\citep{liang2026vip}, dynamic submodels for video segmentation~\citep{tang2026rds}, and acceleration policies inside diffusion transformers~\citep{zhao2026rapid3}. Coordinators, harness-level routers, and token-level routers extend routing beyond model identity~\citep{nielsen2026conductor,xu2026trinity,liu2026harness,fu2025r2r,fumolab2026fusion}. We study routing across output types on one matrix: language models, image, video, and speech generators, and robot policies, all scored on shared inputs.

\paragraph{Portfolios and tools.}
Algorithm portfolios choose among complementary solvers per instance~\citep{rice1976algorithm,gomes2001portfolios,kotthoff2014survey}. Their virtual best solver corresponds to our per-input oracle, and its advantage over the single best solver gives the potential gain in Proposition~\ref{prop:ladder}. Residual coverage measures what an additional solver contributes to the portfolio~\citep{xu2012evaluating,frechette2016shapley}. Accounting for call costs connects this selection problem to fractional programming, which optimizes ratio objectives such as quality per unit cost~\citep{dinkelbach1967nonlinear,charnes1962programming}. The value of information addresses the related question of whether acquiring a routing signal is worth its cost~\citep{raiffa1961applied,howard1966information}. Similar cost considerations arise in test-time compute, where gains from extra calls depend on the task and allocation~\citep{chen2024more,snell2025scaling}, while imperfect verifiers can limit the gains from resampling~\citep{stroebl2024inference}. Feedback-driven self-repair uses earlier attempts to guide later ones~\citep{chen2024selfdebug,shinn2023reflexion}, making resampling at equal cost a natural baseline~\citep{olausson2024selfrepair}.

\section{Conclusion}
\label{sec:conclusion}

Across text, images, video, speech, code, and robots, the best candidate changes from input to input, and the per-input oracle sits 5.4 points above the best fixed candidate per benchmark. Most of that potential gain lies within task types, and a learned type router already scores 0.3 points above the best fixed candidate across 16 benchmarks, and up to 2.1 points within an output group. 
In embodied scenarios, a router trained on camera frames tracks the best VLA policy in each scene condition and beats the best single policy by 1.1 points with 24\% less policy time. Routing's most reliable gain is a cheaper route that is good enough, raising efficiency over the reference by $8.82\times$ on average: when candidates are close in quality and far apart in cost, the reference and the quality tolerance fix the price, and a finite-sample test certifies the quality floor. Three rules follow: \textbf{(1)}~compare candidates on the same inputs, and measure gains against the matched best fixed candidate; \textbf{(2)}~train routers on per-candidate outcomes, and look for gains within task types; \textbf{(3)}~fix the reference and the tolerance before ranking routes, and certify the floor on independent inputs.


\clearpage
\phantomsection
\addcontentsline{toc}{section}{References}
\renewcommand{\bibfont}{\small}
\setlength{\bibsep}{2pt plus 1pt}
\bibliography{custom}

@inproceedings{dekoninck2025unified,
  author = {Dekoninck, Jasper and Baader, Maximilian and Vechev, Martin},
  title = {A Unified Approach to Routing and Cascading for {LLM}s},
  booktitle = {Proceedings of the 42nd International Conference on Machine Learning},
  volume = {267},
  pages = {12987--13010},
  year = {2025}
}

@article{chen2023frugalgpt,
  author = {Chen, Lingjiao and Zaharia, Matei and Zou, James},
  title = {{FrugalGPT}: How to Use Large Language Models While Reducing Cost and Improving Performance},
  journal = {Transactions on Machine Learning Research},
  year = {2024}
}

@inproceedings{ong2024routellm,
  title={{RouteLLM}: Learning to route {LLM}s from preference data},
  author={Ong, Isaac and Almahairi, Amjad and Wu, Vincent and Chiang, Wei-Lin and Wu, Tianhao and Gonzalez, Joseph E and Kadous, Mohammed and Stoica, Ion},
  booktitle={International Conference on Learning Representations},
  volume={2025},
  pages={34433--34448},
  year={2025}
}

@article{zhou2026agent,
  title={{Agent-as-a-Router}: Agentic Model Routing for Coding Tasks},
  author={Zhou, Pengfei and Tang, Zhiwei and Ma, Yixing and Tang, Jiasheng and Han, Yizeng and Wan, Zhenglin and Meng, Fanqing and Wang, Wei and Zhuang, Bohan and Zhao, Wangbo and others},
  journal={arXiv preprint arXiv:2606.22902},
  year={2026}
}

@article{liu2026harness,
  title={Agentic Routing: The Harness-Native Data Flywheel},
  author={Liu, Xinchen and Zhou, Hang and Zong, Yingjie and Tian, Yuchuan and Song, Liuyang and Zhang, Shuo and Li, Yulong and He, Wei and Zheng, Mengyu and Liu, Runke and others},
  journal={arXiv preprint arXiv:2607.11399},
  year={2026}
}

@article{yu2026scope,
  title={{SCOPE-Router}: Cost-Aware Open-Set {VLM} Routing for Execution-Oriented Tasks},
  author={Yu, Tao and Qu, Yifei and Cui, Zhiqing and Zhou, Pengfei and Luo, Zhongtian and Yang, Yujia and Chai, Shenghua and Jin, Haopeng and Zhang, Zhenghao and Wang, Xinming and others},
  journal={arXiv preprint arXiv:2608.12127},
  year={2026}
}

@article{liang2026vip,
  title={Beyond One-Size-Fits-All: Sample-Adaptive Strategy Routing for Vision Token Pruning in {MLLM}s},
  author={Liang, Haiji and Zhou, Pengfei and Wan, Zhenglin and Wang, Wei and You, Yang and Zhao, Wangbo},
  journal={arXiv preprint arXiv:2609.10346},
  year={2026}
}

@inproceedings{tang2026rds,
  title={Efficient Video Object Segmentation and Tracking with Recurrent Dynamic Submodel},
  author={Tang, Weidong and Liang, Zhiyuan and Wan, Xinyan and Zhu, Chen and Xu, Zhaopan and Zhou, Pengfei and Song, Yan and You, Yang and Zhao, Wangbo},
  booktitle={Proceedings of the IEEE/CVF Conference on Computer Vision and Pattern Recognition},
  pages={20912--20921},
  year={2026}
}

@inproceedings{zhao2026rapid3,
  title={{RAPID}$^3$: Tri-Level Reinforced Acceleration Policies for Diffusion Transformer},
  author={Zhao, Wangbo and Han, Yizeng and Tang, Zhiwei and Tang, Jiasheng and Zhou, Pengfei and Wang, Kai and Zhuang, Bohan and Wang, Zhangyang and Wang, Fan and You, Yang},
  booktitle={International Conference on Learning Representations},
  volume={2026},
  pages={149423--149446},
  year={2026}
}

@inproceedings{nielsen2026conductor,
  title={Learning to orchestrate agents in natural language with the conductor},
  author={Nielsen, Stefan and Cetin, Edoardo and Schwendeman, Peter and Sun, Qi and Xu, Jinglue and Tang, Yujin},
  booktitle={International Conference on Learning Representations},
  volume={2026},
  pages={135686--135724},
  year={2026}
}

@inproceedings{xu2026trinity,
  title={Trinity: An evolved {LLM} coordinator},
  author={Xu, Jinglue and Sun, Qi and Schwendeman, Peter and Nielsen, Stefan and Cetin, Edoardo and Tang, Yujin},
  booktitle={International Conference on Learning Representations},
  volume={2026},
  pages={70195--70224},
  year={2026}
}

@misc{fumolab2026fusion,
  author = {Zhou, Shunfan and Long, Jingqi and Zhang, Yiling and Du, Zhiheng},
  title = {Fusion: Allocating Intelligence Across Heterogeneous Models---{A} New Scaling Dimension for {AI} Systems},
  howpublished = {Whitepaper, Rev.~0.1, Fumo Lab},
  year = {2026},
  url = {https://fumolab.ai/downloads/FUMO_Whitepaper.pdf}
}

@incollection{rice1976algorithm,
  title={The algorithm selection problem},
  author={Rice, John R},
  booktitle={Advances in computers},
  volume={15},
  pages={65--118},
  publisher={Academic Press},
  year={1976}
}

@article{gomes2001portfolios,
  title={Algorithm portfolios},
  author={Gomes, Carla P and Selman, Bart},
  journal={Artificial Intelligence},
  volume={126},
  number={1-2},
  pages={43--62},
  year={2001}
}

@incollection{kotthoff2014survey,
  title={Algorithm selection for combinatorial search problems: A survey},
  author={Kotthoff, Lars},
  booktitle={Data mining and constraint programming: Foundations of a cross-disciplinary approach},
  pages={149--190},
  publisher={Springer},
  year={2016}
}

@inproceedings{xu2012evaluating,
  title={Evaluating component solver contributions to portfolio-based algorithm selectors},
  author={Xu, Lin and Hutter, Frank and Hoos, Holger and Leyton-Brown, Kevin},
  booktitle={International conference on theory and applications of satisfiability testing},
  pages={228--241},
  year={2012}
}

@inproceedings{frechette2016shapley,
  title = {Using the {Shapley} Value to Analyze Algorithm Portfolios},
  author={Fr{\'e}chette, Alexandre and Kotthoff, Lars and Michalak, Tomasz and Rahwan, Talal and Hoos, Holger and Leyton-Brown, Kevin},
  booktitle={Proceedings of the AAAI Conference on Artificial Intelligence},
  volume={30},
  year={2016}
}

@book{raiffa1961applied,
  title={Applied Statistical Decision Theory},
  author={Raiffa, Howard and Schlaifer, Robert},
  publisher={Division of Research, Graduate School of Business Administration, Harvard University},
  year={1961}
}

@article{howard1966information,
  title={Information value theory},
  author={Howard, Ronald A},
  journal={IEEE Transactions on Systems Science and Cybernetics},
  volume={2},
  number={1},
  pages={22--26},
  year={1966}
}

@article{chow1970optimum,
  title={On optimum recognition error and reject tradeoff},
  author={Chow, Chao},
  journal={IEEE Transactions on information theory},
  volume={16},
  number={1},
  pages={41--46},
  year={1970}
}

@inproceedings{jitkrittum2023does,
  title={When does confidence-based cascade deferral suffice?},
  author={Jitkrittum, Wittawat and Gupta, Neha and Menon, Aditya K and Narasimhan, Harikrishna and Rawat, Ankit and Kumar, Sanjiv},
  booktitle={Advances in Neural Information Processing Systems},
  volume={36},
  pages={9891--9906},
  year={2023}
}

@inproceedings{gupta2024language,
  title={Language model cascades: Token-level uncertainty and beyond},
  author={Gupta, Neha and Narasimhan, Harikrishna and Jitkrittum, Wittawat and Rawat, Ankit Singh and Menon, Aditya Krishna and Kumar, Sanjiv},
  booktitle={International Conference on Learning Representations},
  volume={2024},
  pages={4147--4180},
  year={2024}
}

@inproceedings{aggarwal2024automix,
  title={{AutoMix}: Automatically mixing language models},
  author={Aggarwal, Pranjal and Madaan, Aman and Anand, Ankit and Potharaju, Srividya Pranavi and Mishra, Swaroop and Zhou, Pei and Gupta, Aditya and Rajagopal, Dheeraj and Kappaganthu, Karthik and Yang, Yiming and others},
  booktitle={Advances in Neural Information Processing Systems},
  volume={37},
  pages={131000--131034},
  year={2024}
}

@inproceedings{chen2020frugalml,
  title={{FrugalML}: How to use {ML} prediction {API}s more accurately and cheaply},
  author={Chen, Lingjiao and Zaharia, Matei and Zou, James},
  booktitle={Advances in neural information processing systems},
  volume={33},
  pages={10685--10696},
  year={2020}
}

@inproceedings{trapeznikov2013supervised,
  title={Supervised sequential classification under budget constraints},
  author={Trapeznikov, Kirill and Saligrama, Venkatesh},
  booktitle={Artificial intelligence and statistics},
  pages={581--589},
  year={2013}
}

@inproceedings{stroebl2024inference,
  title={The limits of inference scaling through resampling},
  author={Stroebl, Benedikt and Kapoor, Sayash and Narayanan, Arvind},
  booktitle={International Conference on Learning Representations},
  volume={2026},
  pages={92601--92633},
  year={2026}
}

@inproceedings{chen2024more,
  title={Are more {LLM} calls all you need? towards the scaling properties of compound {AI} systems},
  author={Chen, Lingjiao and Davis, Jared and Hanin, Boris and Bailis, Peter and Stoica, Ion and Zaharia, Matei and Zou, James},
  booktitle={Advances in Neural Information Processing Systems},
  volume={37},
  pages={45767--45790},
  year={2024}
}

@inproceedings{snell2025scaling,
  title={Scaling {LLM} test-time compute optimally can be more effective than scaling parameters for reasoning},
  author={Snell, Charlie and Lee, Jaehoon and Xu, Kelvin and Kumar, Aviral},
  booktitle={International Conference on Learning Representations},
  volume={2025},
  pages={10131--10165},
  year={2025}
}

@article{dinkelbach1967nonlinear,
  title={On nonlinear fractional programming},
  author={Dinkelbach, Werner},
  journal={Management science},
  volume={13},
  number={7},
  pages={492--498},
  year={1967}
}

@article{hu2024routerbench,
  title={Routerbench: A benchmark for multi-{LLM} routing system},
  author={Hu, Qitian Jason and Bieker, Jacob and Li, Xiuyu and Jiang, Nan and Keigwin, Benjamin and Ranganath, Gaurav and Keutzer, Kurt and Upadhyay, Shriyash Kaustubh},
  journal={arXiv preprint arXiv:2403.12031},
  year={2024}
}

@inproceedings{fu2025r2r,
  title={{R2R}: Efficiently navigating divergent reasoning paths with small-large model token routing},
  author={Fu, Tianyu and Ge, Yi and You, Yichen and Liu, Enshu and Yuan, Zhihang and Dai, Guohao and Yan, Shengen and Yang, Huazhong and Wang, Yu},
  booktitle={Advances in Neural Information Processing Systems},
  volume={38},
  pages={124108--124145},
  year={2026}
}

@article{charnes1962programming,
  title={Programming with linear fractional functionals},
  author={Charnes, Abraham and Cooper, William W},
  journal={Naval Research logistics quarterly},
  volume={9},
  number={3-4},
  pages={181--186},
  year={1962}
}

@article{westfall2001optimally,
  title={Optimally weighted, fixed sequence and gatekeeper multiple testing procedures},
  author={Westfall, Peter H. and Krishen, Alok},
  journal={Journal of Statistical Planning and Inference},
  volume={99},
  number={1},
  pages={25--40},
  year={2001},
  doi={10.1016/S0378-3758(01)00077-5}
}

@article{angelopoulos2025learn,
  title={Learn then test: Calibrating predictive algorithms to achieve risk control},
  author={Angelopoulos, Anastasios N. and Bates, Stephen and Cand{\`e}s, Emmanuel J. and Jordan, Michael I. and Lei, Lihua},
  journal={The Annals of Applied Statistics},
  volume={19},
  number={2},
  pages={1641--1662},
  year={2025},
  doi={10.1214/24-AOAS1998}
}

@article{hoeffding1963probability,
  title={Probability inequalities for sums of bounded random variables},
  author={Hoeffding, Wassily},
  journal={Journal of the American Statistical Association},
  volume={58},
  number={301},
  pages={13--30},
  year={1963},
  doi={10.1080/01621459.1963.10500830}
}

@inproceedings{maurer2009empirical,
  title={Empirical {B}ernstein bounds and sample variance penalization},
  author={Maurer, Andreas and Pontil, Massimiliano},
  booktitle={Proceedings of the 22nd Annual Conference on Learning Theory ({COLT})},
  year={2009}
}

@inproceedings{thomas2015high,
  title={High Confidence Policy Improvement},
  author={Thomas, Philip and Theocharous, Georgios and Ghavamzadeh, Mohammad},
  booktitle={Proceedings of the 32nd International Conference on Machine Learning},
  volume={37},
  pages={2380--2388},
  year={2015}
}

@inproceedings{ding2024hybrid,
  author    = {Dujian Ding and Ankur Mallick and Chi Wang and Robert Sim and Subhabrata Mukherjee and Victor R{\"u}hle and Laks V. S. Lakshmanan and Ahmed Hassan Awadallah},
  title     = {Hybrid {LLM}: Cost-Efficient and Quality-Aware Query Routing},
  booktitle = {The Twelfth International Conference on Learning Representations},
  year      = {2024}
}

@inproceedings{chen2024routerdc,
  author    = {Shuhao Chen and Weisen Jiang and Baijiong Lin and James T. Kwok and Yu Zhang},
  title     = {{RouterDC}: Query-Based Router by Dual Contrastive Learning for Assembling Large Language Models},
  booktitle = {Advances in Neural Information Processing Systems},
  volume    = {37},
  pages     = {66305--66328},
  year      = {2024}
}

@inproceedings{feng2025graphrouter,
  author    = {Tao Feng and Yanzhen Shen and Jiaxuan You},
  title     = {{GraphRouter}: A Graph-based Router for {LLM} Selections},
  booktitle = {The Thirteenth International Conference on Learning Representations},
  year      = {2025}
}

@inproceedings{zhuang2025embedllm,
  author    = {Richard Zhuang and Tianhao Wu and Zhaojin Wen and Andrew Li and Jiantao Jiao and Kannan Ramchandran},
  title     = {{EmbedLLM}: Learning Compact Representations of Large Language Models},
  booktitle = {The Thirteenth International Conference on Learning Representations},
  year      = {2025}
}

@inproceedings{ding2025bestroute,
  author    = {Dujian Ding and Ankur Mallick and Shaokun Zhang and Chi Wang and Daniel Madrigal and Mirian del Carmen Hipolito Garcia and Menglin Xia and Laks V. S. Lakshmanan and Qingyun Wu and Victor R{\"u}hle},
  title     = {{BEST}-Route: Adaptive {LLM} Routing with Test-Time Optimal Compute},
  booktitle = {Proceedings of the 42nd International Conference on Machine Learning},
  volume    = {267},
  pages     = {13870--13884},
  year      = {2025}
}

@inproceedings{chen2024selfdebug,
  author    = {Xinyun Chen and Maxwell Lin and Nathanael Sch{\"a}rli and Denny Zhou},
  title     = {Teaching Large Language Models to Self-Debug},
  booktitle = {The Twelfth International Conference on Learning Representations},
  year      = {2024}
}

@inproceedings{shinn2023reflexion,
  author    = {Noah Shinn and Federico Cassano and Ashwin Gopinath and Karthik Narasimhan and Shunyu Yao},
  title     = {Reflexion: Language Agents with Verbal Reinforcement Learning},
  booktitle = {Advances in Neural Information Processing Systems},
  volume    = {36},
  pages     = {8634--8652},
  year      = {2023}
}

@inproceedings{olausson2024selfrepair,
  author    = {Theo X. Olausson and Jeevana Priya Inala and Chenglong Wang and Jianfeng Gao and Armando Solar-Lezama},
  title     = {Is Self-Repair a Silver Bullet for Code Generation?},
  booktitle = {The Twelfth International Conference on Learning Representations},
  year      = {2024}
}

@inproceedings{wang2024mmlupro,
  title={{MMLU-Pro}: A More Robust and Challenging Multi-Task Language Understanding Benchmark},
  author={Wang, Yubo and Ma, Xueguang and Zhang, Ge and Ni, Yuansheng and Chandra, Abhranil and Guo, Shiguang and Ren, Weiming and Arulraj, Aaran and He, Xuan and Jiang, Ziyan and others},
  booktitle={Advances in Neural Information Processing Systems, Datasets and Benchmarks Track},
  year={2024}
}

@inproceedings{gema2025mmluredux,
  title={Are We Done with {MMLU}?},
  author={Gema, Aryo Pradipta and Leang, Joshua Ong Jun and Hong, Giwon and Devoto, Alessio and Mancino, Alberto Carlo Maria and Saxena, Rohit and He, Xuanli and Zhao, Yu and Du, Xiaotang and Madani, Mohammad Reza Ghasemi and others},
  booktitle={Proceedings of the Conference of the Nations of the Americas Chapter of the Association for Computational Linguistics},
  year={2025}
}

@inproceedings{du2025supergpqa,
  title={{SuperGPQA}: Scaling {LLM} Evaluation across 285 Graduate Disciplines},
  author={{M-A-P Team} and Du, Xinrun and Yao, Yifan and Ma, Kaijing and others},
  booktitle={Advances in Neural Information Processing Systems, Datasets and Benchmarks Track},
  year={2025}
}

@inproceedings{rein2024gpqa,
  title={{GPQA}: A Graduate-Level {Google}-Proof {Q\&A} Benchmark},
  author={Rein, David and Hou, Betty Li and Stickland, Asa Cooper and Petty, Jackson and Pang, Richard Yuanzhe and Dirani, Julien and Michael, Julian and Bowman, Samuel R.},
  booktitle={Conference on Language Modeling},
  year={2024}
}

@misc{aa2026evals,
  title={{Artificial Analysis} Intelligence Benchmarking Methodology},
  author={{Artificial Analysis}},
  year={2026},
  howpublished={\url{https://artificialanalysis.ai/methodology/intelligence-benchmarking}}
}

@article{phan2025hle,
  title={A benchmark of expert-level academic questions to assess {AI} capabilities},
  author={{Center for AI Safety} and {Scale AI} and {HLE Contributors Consortium}},
  journal={Nature},
  volume={649},
  number={8099},
  pages={1139--1146},
  year={2026},
  doi={10.1038/s41586-025-09962-4}
}

@inproceedings{yue2024mmmu,
  title={{MMMU}: A Massive Multi-discipline Multimodal Understanding and Reasoning Benchmark for Expert {AGI}},
  author={Yue, Xiang and Ni, Yuansheng and Zhang, Kai and Zheng, Tianyu and Liu, Ruoqi and Zhang, Ge and Stevens, Samuel and Jiang, Dongfu and Ren, Weiming and Sun, Yuxuan and others},
  booktitle={Proceedings of the IEEE/CVF Conference on Computer Vision and Pattern Recognition},
  year={2024}
}

@inproceedings{yue2025mmmupro,
  title={{MMMU-Pro}: A More Robust Multi-discipline Multimodal Understanding Benchmark},
  author={Yue, Xiang and Zheng, Tianyu and Ni, Yuansheng and Wang, Yubo and Zhang, Kai and Tong, Shengbang and Sun, Yuxuan and Yu, Botao and Zhang, Ge and Sun, Huan and others},
  booktitle={Proceedings of the Annual Meeting of the Association for Computational Linguistics},
  year={2025}
}

@inproceedings{chang2025oneig,
  title={{OneIG-Bench}: Omni-dimensional Nuanced Evaluation for Image Generation},
  author={Chang, Jingjing and Fang, Yixiao and Xing, Peng and Wu, Shuhan and Cheng, Wei and Wang, Rui and Zeng, Xianfang and Yu, Gang and Chen, Hai-Bao},
  booktitle={Advances in Neural Information Processing Systems, Datasets and Benchmarks Track},
  year={2025}
}

@inproceedings{niu2025wise,
  title={{WISE}: World Knowledge-Informed Semantic Evaluation for Text-to-Image Generation},
  author={Niu, Yuwei and Ning, Munan and Zheng, Mengren and others},
  booktitle={Proceedings of the 43rd International Conference on Machine Learning},
  volume={306},
  pages={93570--93590},
  year={2026}
}

@article{zheng2025vbench2,
  title={{VBench-2.0}: Advancing Video Generation Benchmark Suite for Intrinsic Faithfulness},
  author={Zheng, Dian and Huang, Ziqi and Liu, Hongbo and Zou, Kai and He, Yinan and Zhang, Fan and others},
  journal={arXiv preprint arXiv:2503.21755},
  year={2025}
}

@article{anastassiou2024seedtts,
  title={{Seed-TTS}: A Family of High-Quality Versatile Speech Generation Models},
  author={Anastassiou, Philip and Chen, Jiawei and Chen, Jitong and others},
  journal={arXiv preprint arXiv:2406.02430},
  year={2024}
}

@inproceedings{patwardhan2025gdpval,
  title={{GDPval}: Evaluating {AI} Model Performance on Real-World Economically Valuable Tasks},
  author={Patwardhan, Tejal and Dias, Rachel and Proehl, Elizabeth and others},
  booktitle={International Conference on Learning Representations},
  year={2026}
}

@inproceedings{tian2024scicode,
  title={{SciCode}: A Research Coding Benchmark Curated by Scientists},
  author={Tian, Minyang and Gao, Luyu and Zhang, Shizhuo Dylan and others},
  booktitle={Advances in Neural Information Processing Systems, Datasets and Benchmarks Track},
  year={2024}
}

@inproceedings{tbench2025,
  title={{Terminal-Bench}: Benchmarking Agents on Hard, Realistic Tasks in Command Line Interfaces},
  author={Merrill, Mike A. and Shaw, Alexander G. and Carlini, Nicholas and Li, Boxuan and Raj, Harsh and Bercovich, Ivan and others},
  booktitle={International Conference on Learning Representations},
  year={2026}
}

@inproceedings{chen2025robotwin2,
  title={{RoboTwin 2.0}: A Scalable Data Generator and Benchmark with Strong Domain Randomization for Robust Bimanual Robotic Manipulation},
  author={Chen, Tianxing and Chen, Zanxin and Chen, Baijun and others},
  booktitle={Proceedings of the 43rd International Conference on Machine Learning},
  volume={306},
  pages={13673--13699},
  year={2026}
}

@inproceedings{black2025pi05,
  title={$\pi_{0.5}$: a Vision-Language-Action Model with Open-World Generalization},
  author={Black, Kevin and Brown, Noah and Darpinian, James and Dhabalia, Karan and Driess, Danny and others},
  booktitle={Proceedings of the 9th Conference on Robot Learning},
  volume={305},
  pages={17--40},
  year={2025}
}

@article{kamath2025geneval2,
  title={{GenEval 2}: Addressing Benchmark Drift in Text-to-Image Evaluation},
  author={Kamath, Amita and Chang, Kai-Wei and Krishna, Ranjay and Zettlemoyer, Luke and Hu, Yushi and Ghazvininejad, Marjan},
  journal={arXiv preprint arXiv:2512.16853},
  year={2025}
}

@inproceedings{yang2026longt2ibench,
  title={{LongT2IBench}: A Benchmark for Evaluating Long Text-to-Image Generation with Graph-Structured Annotations},
  author={Yang, Zhichao and Gu, Tianjiao and Wang, Jianjie and Lin, Feiyu and Sheng, Xiangfei and Chen, Pengfei and Li, Leida},
  booktitle={Proceedings of the AAAI Conference on Artificial Intelligence},
  volume={40},
  pages={11820--11828},
  year={2026}
}

@article{li2026qwenimagebench,
  title={{Qwen-Image-Bench}: From Generation to Creation in Text-to-Image Evaluation},
  author={Li, Niantong and Hu, Guangzheng and Qiao, Weixu and Ba, Ying and Hong, Qichen and others},
  journal={arXiv preprint arXiv:2605.28091},
  year={2026}
}

@article{jackson2025omniscience,
  title={{AA-Omniscience}: Evaluating Cross-Domain Knowledge Reliability in Large Language Models},
  author={Jackson, Declan and Keating, William and Cameron, George and Hill-Smith, Micah},
  journal={arXiv preprint arXiv:2511.13029},
  year={2025}
}

@article{shepard2026automationbench,
  title={{AutomationBench}},
  author={Shepard, Daniel and Salimans, Robin},
  journal={arXiv preprint arXiv:2604.18934},
  year={2026}
}

@article{garre2026gdppdf,
  title={{GDP.pdf}: Benchmarking Grounded Multimodal Reasoning over Professional {PDF} Documents},
  author={Garre, Suhaas and Ritchie, Emily and Mehta, Sushant and Chen, Edwin},
  journal={arXiv preprint arXiv:2607.11192},
  year={2026}
}

@inproceedings{zheng2026xvla,
  title={{X-VLA}: Soft-Prompted Transformer as Scalable Cross-Embodiment Vision-Language-Action Model},
  author={Zheng, Jinliang and Li, Jianxiong and Wang, Zhihao and Liu, Dongxiu and Kang, Xirui and others},
  booktitle={International Conference on Learning Representations},
  year={2026}
}

@inproceedings{li2026spatialforcing,
  title={{Spatial Forcing}: Implicit Spatial Representation Alignment for Vision-Language-Action Model},
  author={Li, Fuhao and Song, Wenxuan and Zhao, Han and Wang, Jingbo and Ding, Pengxiang and Wang, Donglin and Zeng, Long and Li, Haoang},
  booktitle={International Conference on Learning Representations},
  year={2026}
}

@article{tang2026trinity,
  title={The Trinity: Capability-Driven Recursive Self-Improvement for Multimodal Agents},
  author={Tang, Weidong and Wang, Ruiheng and Zhang, Jingxuan and Wang, Hexin and Chen, Haoyuan and Zhang, Zhengfeiyang and Ma, Yixing and Jiang, Zichen and Guo, Ziyu and Li, Haodong and Jiao, Zhengbo and Cui, Zhiqing and Yin, Xinxiang and Liang, Shuo and Wan, Zhenglin and Lin, Hongzhan and Zhao, Wangbo and Zhang, Kaipeng and Zhou, Pengfei and Zhang, Odin and You, Yang},
  journal={arXiv preprint arXiv:2610.36077},
  year={2026},
  doi={10.48550/arXiv.2610.36077},
  url={https://arxiv.org/abs/2610.36077}
}

\newpage
\appendix
\onecolumn
\section{Proofs and Formal Results}
\label{app:proofs}
This appendix proves the results of Section~\ref{sec:model} and Corollary~\ref{cor:admission}, and states, with short proofs, the formal results behind Sections~\ref{sec:matched} and~\ref{sec:spend-more}. Except in the proofs of Propositions~\ref{prop:ladder} and~\ref{prop:targets}, where $x$ is the input as in Section~\ref{sec:model}, we write $x$ for the router's observed state $s$ and $\mathcal X$ for its finite set of values.

\theoremstyle{plain}
\newtheorem*{restatedprop}{\restatedpropname}
\newcommand{\restatedpropname}{Proposition}

\subsection{Proof of Lemma~\ref{thm:surplus}}
\label{app:proof-surplus}
\begin{proof}
Write $V=V_g$, $D=D_g$, $\bar v=\E[Y_B]$, and $\bar c=\E[C_B]$. The definitions give $\E[V]=\bar v+Q$ and $\E[D]=\bar c-\mathcal S$, so
\[
\eta-\eta_B=\frac{\bar v+Q}{\bar c-\mathcal S}-\frac{\bar v}{\bar c}
=\frac{Q+(\bar v/\bar c)\mathcal S}{\bar c-\mathcal S},
\]
whose denominator $\E[D]$ is positive. Substituting $\bar v=\eta_B\bar c$ gives, for every $\lambda$,
\begin{equation}
Q+\eta_B\mathcal S=\E[V-\eta_BD],\qquad\E[V-\lambda D]=\E[D]\,(\eta-\lambda).
\label{eq:unified-identity}
\end{equation}
No independence or stationarity is needed; the quality condition follows separately from $\E[V]=\bar v+Q$. For a stationary ergodic sequence of rounds, the long-run ratio $\sum_{n<N}V_n/\sum_{n<N}D_n$ converges almost surely to $\E[V]/\E[D]$, a ratio of aggregates rather than the average of $V_n/D_n$.
\end{proof}

\subsection{Proof of Theorem~\ref{thm:optimal-routing}}
\label{app:proof-optimal-routing}
Throughout, $x$ ranges over a finite set $\mathcal X$ of observable states with router-independent probabilities $\pi(x)>0$; each state offers a finite menu $\mathcal U(x)$ of candidates with expected value $v(x,u)$ and expected cost $c(x,u)>0$; a router $g$ chooses $u$ with probability $g(u\mid x)$. Let $\mathcal G$ be the set of routers, a product of simplices and hence a compact polytope, and put $c_{\min}=\min_{x,u}c(x,u)>0$. For $g\in\mathcal G$,
\[
\E_g[V]=\sum_x\pi(x)\sum_{u\in\mathcal U(x)}g(u\mid x)\,v(x,u),\qquad
\E_g[D]=\sum_x\pi(x)\sum_{u\in\mathcal U(x)}g(u\mid x)\,c(x,u)\ \ge\ c_{\min},
\]
$\eta(g)=\E_g[V]/\E_g[D]$, and for every real $\lambda$
\begin{equation}
\E_g[V-\lambda D]=\E_g[D]\,\bigl(\eta(g)-\lambda\bigr).
\label{eq:app-dinkelbach-identity}
\end{equation}
A router is deterministic if every $g(\cdot\mid x)$ is a point mass. The reference policy is admissible if it is an element $g_B\in\mathcal G$; then $\eta(g_B)=\eta_B$ and $\E_{g_B}[V]=\bar v$, so the reference satisfies the floor $\E[V]\ge\bar v-\epsilon$ for every $\epsilon\ge0$.

\begin{proof}
\emph{Two preliminary facts.}
(F1) State-wise separation: for any real numbers $\phi(x,u)$, the linear functional $g\mapsto\sum_x\pi(x)\sum_ug(u\mid x)\phi(x,u)$ on $\mathcal G$ is maximized exactly by the routers that, in every state $x$, put all their mass on $\argmax_u\phi(x,u)$, and its maximum is $\sum_x\pi(x)\max_u\phi(x,u)$. Indeed, for each $x$ the inner sum is a convex combination of the numbers $\phi(x,u)$, hence at most $\max_u\phi(x,u)$, with equality if and only if $g(\cdot\mid x)$ is supported on the maximizers; and $\pi(x)>0$. In particular a deterministic maximizer exists.
(F2) Sign of the Dinkelbach function: applying (F1) with $\phi=v-\lambda c$,
\begin{equation}
\Phi(\lambda)=\sum_x\pi(x)\max_{u\in\mathcal U(x)}\bigl[v(x,u)-\lambda c(x,u)\bigr]=\max_{g\in\mathcal G}\E_g[V-\lambda D],
\label{eq:app-phi-max}
\end{equation}
so by \eqref{eq:app-dinkelbach-identity} and $\E_g[D]>0$, $\Phi(\lambda)\ge0$ if and only if some router has $\eta(g)\ge\lambda$, and $\Phi(\lambda)>0$ if and only if some router has $\eta(g)>\lambda$.

\smallskip\noindent\emph{Part (i).}
Each $\phi_x(\lambda)=\max_u[v(x,u)-\lambda c(x,u)]$ is the maximum of finitely many affine functions of $\lambda$ with slopes $-c(x,u)<0$. It is therefore convex, piecewise linear and continuous, and strictly decreasing: if $\lambda<\lambda'$ and $u'$ attains $\phi_x(\lambda')$, then $\phi_x(\lambda)\ge v(x,u')-\lambda c(x,u')>v(x,u')-\lambda'c(x,u')=\phi_x(\lambda')$. Moreover $\phi_x(\lambda)\to+\infty$ as $\lambda\to-\infty$, and $\phi_x(\lambda)\le\max_uv(x,u)-\lambda c_{\min}\to-\infty$ as $\lambda\to+\infty$. The positive combination $\Phi=\sum_x\pi(x)\phi_x$ inherits all these properties and has a unique root $\eta^\star$. By (F2), $\Phi(\eta^\star)=0$ yields a router with $\eta(g)\ge\eta^\star$ and forbids $\eta(g)>\eta^\star$; hence $\eta^\star=\max_g\eta(g)$, and the maximum is attained. A router $g$ is efficiency-optimal if and only if $\eta(g)=\eta^\star$, if and only if $\E_g[V-\eta^\star D]=0=\Phi(\eta^\star)$ by \eqref{eq:app-dinkelbach-identity}, if and only if $g$ maximizes the $\eta^\star$-objective over $\mathcal G$, if and only if, by (F1), in every state it uses only candidates maximizing $v(x,u)-\eta^\star c(x,u)$.

\smallskip\noindent\emph{Part (ii): existence.}
Let $\mathcal G_\epsilon=\{g\in\mathcal G\colon\E_g[V]\ge\bar v-\epsilon\}$, a compact polytope that contains $g_B$ because $\epsilon\ge0$. Since $\eta$ is continuous on $\mathcal G$ (its denominator is at least $c_{\min}$), it attains its maximum $\eta^\star_\epsilon$ on $\mathcal G_\epsilon$.

\smallskip\noindent\emph{Part (ii): the multiplier.}
Define $\Phi_\epsilon(\lambda)=\max_{g\in\mathcal G_\epsilon}\E_g[V-\lambda D]$. As a maximum over a compact set of affine functions of $\lambda$ with slopes $-\E_g[D]\le-c_{\min}$, it is convex, continuous and strictly decreasing, tends to $\pm\infty$ as $\lambda\to\mp\infty$, and has a unique root. By \eqref{eq:app-dinkelbach-identity}, $\Phi_\epsilon(\lambda)\ge0$ if and only if some $g\in\mathcal G_\epsilon$ has $\eta(g)\ge\lambda$, so the root is $\eta^\star_\epsilon$. Consequently a router is constrained-optimal if and only if it solves
\begin{equation}
\max_{g\in\mathcal G}\ f(g):=\E_g[V-\eta^\star_\epsilon D]
\quad\text{subject to}\quad
h(g):=\E_g[V]-(\bar v-\epsilon)\ge0,
\label{eq:app-floor-lp}
\end{equation}
whose optimal value is $\Phi_\epsilon(\eta^\star_\epsilon)=0$. Both $f$ and $h$ are affine in $g$ and $\mathcal G$ is a polytope, so \eqref{eq:app-floor-lp} is a linear program; it is feasible ($g_B$) and bounded. By linear-programming strong duality applied to the single constraint $h\ge0$, there is a multiplier $\kappa\ge0$ such that
\begin{equation*}
\max_{g\in\mathcal G}\,\bigl[f(g)+\kappa\,h(g)\bigr]=0 .
\end{equation*}
If $g$ is optimal for \eqref{eq:app-floor-lp}, then $h(g)\ge0$ and $f(g)=0$, so $0\le f(g)+\kappa h(g)\le0$: hence $\kappa h(g)=0$ and $g$ maximizes $f+\kappa h$ over $\mathcal G$. Now $f(g)+\kappa h(g)=\E_g[(1+\kappa)V-\eta^\star_\epsilon D]+\kappa(\epsilon-\bar v)$, so by (F1) every constrained-optimal router uses, in every state, only candidates maximizing $(1+\kappa)v(x,u)-\eta^\star_\epsilon c(x,u)$, that is, $v(x,u)-\lambda_\epsilon c(x,u)$ with $\lambda_\epsilon=\eta^\star_\epsilon/(1+\kappa)$, and so maximizes $J_{\lambda_\epsilon}$; and $\kappa\,(\E_g[V]-\bar v+\epsilon)=0$: if $\kappa>0$ the floor binds at every optimal router. The value $\kappa=0$ is a valid multiplier if and only if some efficiency-optimal router of part~(i) satisfies the floor, and then $\eta^\star_\epsilon=\eta^\star$.
\end{proof}

\subsection{Proof of Proposition~\ref{prop:ladder}}
\label{app:proof-ladder}
\begin{proof}
Assume each $W_u$ is integrable, and write $S_1=\sigma_1(x)$ and $S_2=\sigma_2(x)$. The menu $\mathcal U(S_1)$ is finite and available at every finer state. By the tower property, $\E[W_u\mid S_1]=\E\bigl[\E[W_u\mid S_2]\bigm|S_1\bigr]\le\E\bigl[\max_{u'}\E[W_{u'}\mid S_2]\bigm|S_1\bigr]$ for every $u$; taking the maximum over $u$ and then expectations gives $V^\star(\sigma_1)\le V^\star(\sigma_2)$. Let $u_1$ attain $\max_u\E[W_u\mid S_1]$. The two sides are equal almost surely if and only if $\E[W_{u_1}\mid S_2]=\max_u\E[W_u\mid S_2]$ almost surely given $S_1$, that is, if and only if one candidate is optimal at almost every finer state; conversely, if a candidate $u$ is optimal at almost every $S_2$ with $\sigma_1=s_1$, then $\E[W_u\mid S_1]$ attains the right side, so equality holds. The same argument with $\sigma_2$ replaced by the identity gives $V^\star(\sigma_2)\le V^\star(x)$. Finally, $W_u\le\max_{u'}W_{u'}$ for every $u$, so $\max_u\E[W_u\mid x]\le\E[\max_uW_u\mid x]$ for every $x$, and taking expectations gives the last inequality; the gap is the expectation of this nonnegative difference. It vanishes if and only if the difference is zero for almost every $x$. If $u(x)$ attains $\max_u\E[W_u\mid x]$, the difference is $\E[\max_uW_u-W_{u(x)}\mid x]$, the mean of a nonnegative variable, which is zero if and only if $W_{u(x)}=\max_uW_u$ almost surely given $x$; and if some candidate attains the maximum almost surely given $x$, it also attains $\max_u\E[W_u\mid x]$.
\end{proof}

\subsection{Proof of Proposition~\ref{prop:targets}}
\label{app:proof-targets}
\begin{proof}
(a)~Fix $x$ and a candidate $u$, and let $\mu=\E[V_u\mid x]\in[0,1]$. For a score $m\in[0,1]$, with $0\log0=0$, the expected cross-entropy is $-\mu\log m-(1-\mu)\log(1-m)$, a strictly convex function of $m$ minimized at $m=\mu$, and the expected squared error is $\operatorname{Var}(V_u\mid x)+(\mu-m)^2$, also minimized at $m=\mu$. Hence $m_u(x)=\E[V_u\mid x]$, and routing to $\argmax_um_u(x)$ earns $\max_u\E[V_u\mid x]$ at every $x$, whose expectation is $V^\star(x)$ at $\lambda=0$.
(b)~Since $\tfrac12<b<a<1$, candidate $B$ beats $A$ exactly when $V_B=1$, an event of probability $b$, and loses otherwise; no ties occur. So $p_B=b>\tfrac12>1-b=p_A$, and the winner-label router plays $B$, whose expected value $b$ is below $a$. The loss $a-b$ approaches $\tfrac12$ as $b\downarrow\tfrac12$ and $a\uparrow1$. In the tie-break example, the first candidate wins only when it succeeds and the second fails, with probability $0.9\times0.5=0.45$.
\end{proof}

\subsection{Proof of Proposition~\ref{prop:certified-floor}}
\label{app:proof-certified-floor}
\paragraph{Setting.}
Let $P$ be the deployment law of a task. A task has a state $x$ in a finite set $\mathcal X$, a menu $\mathcal U(x)$, a realized value $V_u\in[0,1]$ and cost $D_u>0$ for every $u\in\mathcal U(x)$, and the reference value $Y_B\in[0,1]$ and cost $C_B$. A router $g$, possibly randomized, is scored by its expected delivered value given the task, $V_g=\sum_ug(u\mid x)V_u$, which is $V_{g(x)}$ for a deterministic router, and $Q_g=\E_P[V_g-Y_B]$ is its quality change. The reference is admissible: it is a router $g_B$ with $\E_P[V_{g_B}]=\E_P[Y_B]$, so $Q_{g_B}=0$. Two samples are independent: $\mathcal A$ is arbitrary fitting data, and $\mathcal B=(T_1,\dots,T_n)$ consists of $n$ i.i.d.\ tasks from $P$, each recording $x_i$, $Y_{B,i}$ and $V_{u,i}$ for every candidate used below, as in matched replay.

From $\mathcal A$ we fit $\hat\pi(x)>0$, $\hat v(x,u)\in[0,1]$ and $\hat c(x,u)>0$. For a router $g$ write $\hat\E_g[V]=\sum_x\hat\pi(x)\sum_ug(u\mid x)\hat v(x,u)$, define $\hat\E_g[D]$ likewise, and put $\hat\eta(g)=\hat\E_g[V]/\hat\E_g[D]$ and $\hat J_\lambda(g)=\hat\E_g[V]-\lambda\hat\E_g[D]$. Let $\hat\phi_x(\lambda)=\max_{u\in\mathcal U(x)}[\hat v(x,u)-\lambda\hat c(x,u)]$ and $\hat\Phi=\sum_x\hat\pi(x)\hat\phi_x$. Theorem~\ref{thm:optimal-routing}(i), applied to the fitted primitives, gives the root $\hat\eta^\star=\max_g\hat\eta(g)$, and we assume $\hat\E_{g_B}[V]>0$, so that $\hat\eta^\star>0$. A \emph{breakpoint} of $\hat\phi_x$ is a point at which $\hat\phi_x$ is not differentiable. Let $\lambda_1<\dots<\lambda_{K-1}$ be the breakpoints of all $\hat\phi_x$ that lie in $(0,\hat\eta^\star)$, and set $\lambda_0=0$, $\lambda_K=\hat\eta^\star$ and $I_k=(\lambda_{k-1},\lambda_k)$. Fix a strict order on each $\mathcal U(x)$ (the tie-break), and let $g_k(x)$ be the first maximizer of $\hat v(x,u)-\lambda\hat c(x,u)$ in that order, for $\lambda\in I_k$. The sequence $g_1,\dots,g_K$ is the \emph{price path}. A mixture $wg_k+(1-w)g_{k+1}$ is the router that, in each state, plays $g_k(x)$ with probability $w$ and $g_{k+1}(x)$ otherwise.

\renewcommand{\restatedpropname}{Proposition~\ref{prop:certified-floor} (full statement)}
\begin{restatedprop}
\emph{(a) Path.} (1)~Each $g_k$ is well defined, is the same for every $\lambda\in I_k$, and maximizes $\hat J_\lambda$ over all routers for every $\lambda\in[\lambda_{k-1},\lambda_k]$. Moreover $K\le1+\sum_x(|\mathcal U(x)|-1)$. (2)~For $k<l$, $\hat\E_{g_k}[V]>\hat\E_{g_l}[V]$, $\hat\E_{g_k}[D]>\hat\E_{g_l}[D]$ and $\hat\eta(g_k)<\hat\eta(g_l)$; in particular $g_1,\dots,g_K$ are distinct.

\emph{(b) Certification.} Let $\alpha\in(0,1)$ and $\epsilon\ge0$, and let $p_1,\dots,p_K$ be functions of $(\mathcal A,\mathcal B)$ such that $\mathbb P(p_k\le\alpha\mid\mathcal A)\le\alpha$ whenever $Q_{g_k}<-\epsilon$. Let $\hat k=\max\{k\in\{0,\dots,K\}:p_j\le\alpha\text{ for all }j\le k\}$, and let $\hat g=g_{\hat k}$ if $\hat k\ge1$ and $\hat g=g_B$ otherwise. Then
\[
\mathbb P\bigl(Q_{\hat g}<-\epsilon\mid\mathcal A\bigr)\le\alpha\quad\text{almost surely, hence}\quad\mathbb P\bigl(Q_{\hat g}<-\epsilon\bigr)\le\alpha .
\]
With $\hat Q_k=n^{-1}\sum_{i=1}^n(V_{g_k,i}-Y_{B,i})$, the Hoeffding $p$-value $p_k=\exp\{-n(\hat Q_k+\epsilon)_+^2/2\}$ satisfies the requirement. The conclusion holds for any finite family of routers and any order that are fixed given $\mathcal A$.
\end{restatedprop}

\begin{proof}
\emph{Part (a)(1).} Fix $x$. As a maximum of $|\mathcal U(x)|$ affine functions with slopes $-\hat c(x,u)<0$, $\hat\phi_x$ is convex and piecewise linear; its slope takes values in $\{-\hat c(x,u):u\in\mathcal U(x)\}$ and strictly increases at each breakpoint, so $\hat\phi_x$ has at most $|\mathcal U(x)|-1$ breakpoints. A breakpoint shared by several states is counted once, so $K-1\le\sum_x(|\mathcal U(x)|-1)$. On $I_k$ no $\hat\phi_x$ has a breakpoint, so $\hat\phi_x(\lambda)=a_x-b_x\lambda$ there. For each $u$, $\hat\phi_x(\lambda)-[\hat v(x,u)-\lambda\hat c(x,u)]$ is affine and nonnegative on $I_k$, so if it vanishes at one point of the open interval $I_k$, it vanishes on all of it. Hence the maximizer set, and its first element $g_k(x)$, are the same for every $\lambda\in I_k$, and every maximizer has $(\hat v(x,u),\hat c(x,u))=(a_x,b_x)$. By fact~(F1) of Appendix~\ref{app:proof-optimal-routing}, $g_k$ maximizes $\hat J_\lambda$ on $I_k$; since $\hat J_\lambda(g_k)$ and $\hat\Phi(\lambda)$ are continuous in $\lambda$ and agree on $I_k$, they agree on its closure.

\emph{Part (a)(2).} Fix $1\le k\le K-1$. Let $\mathcal X_k$ be the set of states for which $\lambda_k$ is a breakpoint; it is nonempty by the definition of $\lambda_k$. For $x\notin\mathcal X_k$, $\hat\phi_x$ is affine on $I_k\cup\{\lambda_k\}\cup I_{k+1}$, so by (a)(1) the candidates $g_k(x)$ and $g_{k+1}(x)$ have the same pair $(\hat v,\hat c)$. For $x\in\mathcal X_k$, $\hat c(x,g_k(x))=-\hat\phi_x'(\lambda_k-)>-\hat\phi_x'(\lambda_k+)=\hat c(x,g_{k+1}(x))$, and both candidates attain $\hat\phi_x(\lambda_k)$ by continuity, so $\hat v(x,g_{k+1}(x))-\hat v(x,g_k(x))=\lambda_k[\hat c(x,g_{k+1}(x))-\hat c(x,g_k(x))]$. Write $(v_j,d_j)=(\hat\E_{g_j}[V],\hat\E_{g_j}[D])$. Because $\hat\pi>0$ and $\lambda_k>0$,
\[
d_{k+1}-d_k=\sum_{x\in\mathcal X_k}\hat\pi(x)\bigl[\hat c(x,g_{k+1}(x))-\hat c(x,g_k(x))\bigr]<0,\qquad v_{k+1}-v_k=\lambda_k(d_{k+1}-d_k)<0 .
\]
Chaining consecutive steps gives $v_k>v_l$ and $d_k>d_l$ for $k<l$. For efficiency, pick $\lambda'\in I_l$. Optimality of $g_l$ at $\lambda'$ gives $v_k-v_l\le\lambda'(d_k-d_l)$, hence
\[
v_ld_k-v_kd_l=v_l(d_k-d_l)-d_l(v_k-v_l)\ge(d_k-d_l)(v_l-\lambda'd_l)=(d_k-d_l)\,\hat\Phi(\lambda')>0,
\]
because $d_k>d_l$ and $\hat\Phi$ is strictly decreasing with root $\hat\eta^\star>\lambda'$ (Theorem~\ref{thm:optimal-routing}(i)). Dividing by $d_kd_l>0$ gives $\hat\eta(g_l)>\hat\eta(g_k)$. Distinct fitted costs make the routers distinct.

\emph{Part (b).} Condition on $\mathcal A$. Then $g_1,\dots,g_K$, their order, and the numbers $Q_{g_k}$ are fixed. Let $k_0=\min\{k:Q_{g_k}<-\epsilon\}$, with $k_0=\infty$ if no path router violates the floor. The reference cannot violate it, since $Q_{g_B}=0\ge-\epsilon$; this is the only use of $\epsilon\ge0$. Suppose $Q_{\hat g}<-\epsilon$. Then $\hat k\ge1$ and $g_{\hat k}$ violates the floor, so $k_0\le\hat k<\infty$. By the definition of $\hat k$, every $p_j$ with $j\le\hat k$ is at most $\alpha$, and in particular $p_{k_0}\le\alpha$. Therefore
\[
\mathbb P\bigl(Q_{\hat g}<-\epsilon\mid\mathcal A\bigr)\le\1\{k_0<\infty\}\,\mathbb P\bigl(p_{k_0}\le\alpha\mid\mathcal A\bigr)\le\alpha ,
\]
by the requirement on $p_{k_0}$, because $Q_{g_{k_0}}<-\epsilon$. Taking expectations over $\mathcal A$ gives the unconditional bound. No condition is placed on the dependence among $p_1,\dots,p_K$, which share the sample $\mathcal B$.

For the Hoeffding $p$-value, condition on $\mathcal A$. Since $\mathcal B$ is independent of $\mathcal A$ and $g_k$ is fixed, the variables $Z_i=V_{g_k,i}-Y_{B,i}$ are i.i.d., take values in $[-1,1]$, and have mean $Q_{g_k}$. Put $t_\alpha=\sqrt{2\log(1/\alpha)/n}>0$. Then $p_k\le\alpha$ if and only if $\hat Q_k+\epsilon\ge t_\alpha$. If $Q_{g_k}<-\epsilon$, Hoeffding's inequality for a range of length two \citep{hoeffding1963probability} gives
\[
\mathbb P(p_k\le\alpha\mid\mathcal A)\le\mathbb P\bigl(\hat Q_k-Q_{g_k}\ge t_\alpha\mid\mathcal A\bigr)\le\exp(-nt_\alpha^2/2)=\alpha .
\qedhere
\]
\end{proof}

\paragraph{Other $p$-values, refined paths, and the certified price.}
Validity does not use the monotonicity in part~(a); the order matters only for power, and part~(a)(2) puts routers with high fitted value, and therefore a small risk of violation, first. Any other valid $p$-value can be used. The empirical-Bernstein bound of \citet[Theorem~4]{maurer2009empirical}, applied to $(1-Z_i)/2$ with $Z_i=V_{g_k,i}-Y_{B,i}$, yields a valid $p$-value whose margin shrinks with the sample variance of $V-Y_B$. Because validity needs only a finite family in an order fixed by $\mathcal A$, mixtures $wg_k+(1-w)g_{k+1}$ with decreasing $w$ may be inserted between consecutive path routers; we insert $w\in\{0.9,0.8,\dots,0.1\}$ and, at each tolerance studied, the fitted floor optimum of Theorem~\ref{thm:optimal-routing}(ii) computed on $\mathcal A$, which on our menus mixes two neighbouring path routers. If $\hat k\ge1$, the certified router maximizes $\hat J_\lambda$ at $\hat\lambda=\lambda_{\hat k}$, the right end of its interval on the path; we call $\hat\lambda$ the \emph{certified price}. It turns the choice of $\lambda$ into an output of the reference, the tolerance $\epsilon$, and the level $\alpha$.

\subsection{Proof of Corollary~\ref{cor:admission}}
\label{app:proof-admission}
A candidate \emph{calls the reference} in state $x$ if it invokes the reference with positive probability there. For a candidate $u$, the set $\Lambda_x(u)$ of prices $\lambda\ge0$ at which $u$ maximizes $v(x,\cdot)-\lambda c(x,\cdot)$ is a closed interval, possibly empty, because $\lambda\mapsto v(x,u)-\lambda c(x,u)-\phi_x(\lambda)$ is concave, continuous, and nonpositive, and $\Lambda_x(u)$ is the set where it vanishes. The admission price $\lambda^B$ is the largest point of the finite union of the sets $\Lambda_x(u)$ over states and reference-calling candidates ($0$ if the union is empty, $+\infty$ if it is unbounded).
\begin{proof}
By Theorem~\ref{thm:optimal-routing}(i), an efficiency-optimal router plays in every state only maximizers of $v(x,\cdot)-\eta^\star c(x,\cdot)$; since $\eta^\star>\lambda^B$, none of them calls the reference. By part~(ii), a floor-optimal router maximizes $J_{\lambda_\epsilon}$, so by part~(i) it plays only maximizers of $v(x,\cdot)-\lambda_\epsilon c(x,\cdot)$, none of which calls the reference when $\lambda_\epsilon>\lambda^B$. For the cascade $u$ with primary $P$, let $P$ also denote direct dispatch to it, with value $Y_P$ and cost $C_P\le A$. By Eq.~\eqref{eq:accounting-main},
\[
v(x,u)-v(x,P)=\E[(Y_B-Y_P)\1_{H=0}\mid x]=:N,\qquad c(x,u)-c(x,P)=\E[A-C_P\mid x]+r\ \ge\ r,
\]
with $r=\E[C_B\1_{H=0}\mid x]>0$. If $u$ maximizes at $\lambda\ge0$, then $v(x,u)-\lambda c(x,u)\ge v(x,P)-\lambda c(x,P)$, so $\lambda\,[c(x,u)-c(x,P)]\le N$. Hence $N\ge0$ and $\lambda\le N/r$, the bound stated. It holds for every verifier and verifier cost, but it needs direct dispatch to $P$ on the menu: with $Y_P=0$, $A=C_P=1$, $Y_B=C_B=1$, $H=0$, and a single other candidate with cost 10 and value 0, the cascade maximizes at every price although $N/r=1$.
\end{proof}

\subsection{Portfolio Comparisons}
\label{app:portfolio}
For a portfolio $S$ and a candidate $m$, let $S^+=S\cup\{m\}$, with delivered values $V_S,V_{S^+}$ and complete costs $D_S,D_{S^+}$ of positive mean on the same tasks. Lemma~\ref{thm:surplus}, with $S$ as the reference, gives the paired gain
\begin{equation}
G_m(S)=\E[V_{S^+}-V_S]-\eta(S)\E[D_{S^+}-D_S],\qquad \eta(S^+)-\eta(S)=\frac{G_m(S)}{\E[D_{S^+}]},
\label{eq:main-paired}
\end{equation}
so adding $m$ improves efficiency if and only if $G_m(S)>0$, whatever it changes in accepted quality, fallback composition, or verifier behaviour. Let $E_m$ be the event that $m$ supplies a sufficient output, and let an exact selecting verifier pick a sufficient member whenever one exists, else call the reference. Then $S$ hits with probability $p(S)=\mathbb P(\bigcup_{i\in S}E_i)$, and the \emph{residual coverage} of $m$ is $r_m(S)=\mathbb P(E_m\cap\bigcap_{i\in S}E_i^c)=p(S^+)-p(S)$, a joint property of the members' failures that standalone success rates do not determine.

\begin{proposition}[Residual coverage and its correction]
\label{thm:marginal}
Let $S$ deliver mean value $v_h$ on a hit and $v_{\rm miss}$ on a miss, at primary cost $a(S)$ plus a fallback cost $b$ on a miss, and let $d_m(S)=a(S^+)-a(S)$. If $S^+$ has the same $v_h$, $v_{\rm miss}$, and $b$, then $G_m(S)$ equals the efficiency-aware residual coverage
\begin{equation}
N_m(S)=r_m(S)\,[v_h-v_{\rm miss}+\eta(S)b]-\eta(S)d_m(S).
\label{eq:main-residual}
\end{equation}
If instead $S^+$ shifts these conditional means by $\delta v_h$, $\delta v_{\rm miss}$, and $\delta b$, then, with $p=p(S)$ and $r=r_m(S)$,
\begin{equation}
G_m(S)=N_m(S)+(p+r)\,\delta v_h+(1-p-r)\bigl[\delta v_{\rm miss}-\eta(S)\,\delta b\bigr].
\label{eq:portfolio-shift-correction}
\end{equation}
\end{proposition}
\begin{proof}
$S^+$ hits with probability $p+r$. By total expectation, $\E[V_{S^+}-V_S]=r(v_h-v_{\rm miss})+(p+r)\delta v_h+(1-p-r)\delta v_{\rm miss}$ and $\E[D_{S^+}-D_S]=d_m(S)-rb+(1-p-r)\delta b$. Substituting into Eq.~\eqref{eq:main-paired} gives Eq.~\eqref{eq:portfolio-shift-correction}, and with no shifts $G_m(S)=N_m(S)$.
\end{proof}

\subsection{Retries and the Value of Routing Information}
\label{app:escalation}
\paragraph{Planning retries.}
A plan that retries after failures is a candidate, so Theorem~\ref{thm:optimal-routing} chooses among plans; a plan's value is computed backward over post-failure histories. A history $\mathfrak h$ records the attempts so far and their feedback. For a next attempt $u$, let $a(\mathfrak h,u)$ be its incremental cost, including verification and its longer context, $p_\theta(\mathfrak h,u)$ its acceptance probability, and $v_{\rm acc}(\mathfrak h,u)$ its accepted value, all conditional on $\mathfrak h$, and let $J^{\rm direct}(\mathfrak h)=\E[Y_B-\lambda C_B\mid\mathfrak h]$ be the return of calling the reference now.
\begin{proposition}[Planning on post-failure states]
\label{prop:attempts}
The optimal remaining return $\Pi$ equals $J^{\rm direct}$ at the horizon and satisfies
\begin{equation}
\Pi(\mathfrak h)=\max\Bigl\{J^{\rm direct}(\mathfrak h),\ \max_u\bigl[-\lambda a(\mathfrak h,u)+p_\theta(\mathfrak h,u)v_{\rm acc}(\mathfrak h,u)+(1-p_\theta(\mathfrak h,u))\,\E[\Pi(\mathfrak h')\mid\mathfrak h,u,H_\theta=0]\bigr]\Bigr\},
\label{eq:main-chain-history}
\end{equation}
where $\mathfrak h'$ extends $\mathfrak h$ by the rejected attempt. An attempt is optimal if and only if it attains the maximum, so its success probability enters only as conditioned on the history that reaches it.
\end{proposition}
\begin{proof}
Stopping returns $J^{\rm direct}(\mathfrak h)$. Attempting $u$ costs $a(\mathfrak h,u)$; with probability $p_\theta(\mathfrak h,u)$ the task ends with value $v_{\rm acc}(\mathfrak h,u)$, and otherwise it continues from $\mathfrak h'$, whose best continuation is worth $\Pi(\mathfrak h')$. Since $J_\lambda$ is linear in value and cost, attempting and then continuing optimally is worth the bracket, a randomized choice cannot beat the better branch, and backward induction from the horizon gives Eq.~\eqref{eq:main-chain-history}.
\end{proof}
Planning on marginal success rates replaces $p_\theta(\mathfrak h,u)$ by the unconditional rate of $u$, which overstates later attempts when failures are positively correlated.

\paragraph{Buying routing information.}
Let accepted outputs and the fallback preserve value, let the primary be sufficient with probability $s$, and let $0<a<b$ be the costs of the primary and of the fallback. Without information the router incurs the lower of $a+(1-s)b$ (attempt, then escalate on failure) and $b$ (direct fallback). A signal $Z$ that moves the posterior to $s_Z$ saves, before its price,
\begin{equation}
\min\{a+(1-s)b,\,b\}-\E\min\{a+(1-s_Z)b,\,b\},
\label{eq:main-signal}
\end{equation}
and its \emph{capture} is this saving divided by the bound below.
\begin{proposition}[The value of routing information]
\label{prop:voi}
For every signal $Z$, the saving~\eqref{eq:main-signal} is at most $\min\{s(b-a),a(1-s)\}$, which exact revelation of the sufficiency event attains. A signal priced at $k$ is therefore worth buying only if $k<a(1-s)\le a$: it avoids at most one unsuccessful attempt per task, whatever its accuracy.
\end{proposition}
\begin{proof}
Put $f(t)=\min\{a+(1-t)b,b\}$ and $h(t)=ta+(1-t)b$. As a minimum of two affine functions, $f$ is concave, and $f\ge h$ on $[0,1]$, since $f-h$ is the smaller of $a(1-t)$ and $t(b-a)$. Since $h$ is affine and $\E[s_Z]=s$, $h(s)=\E[h(s_Z)]\le\E[f(s_Z)]$, so the saving $f(s)-\E[f(s_Z)]$ is at most $f(s)-h(s)=\min\{s(b-a),a(1-s)\}$. Exact revelation has $s_Z\in\{0,1\}$, where $f=h$, so it attains the bound; this is the expected value of perfect information for two candidates and a binary state~\citep{raiffa1961applied,howard1966information}.
\end{proof}

\subsection{Router Memory}
\label{app:memory}
A record store refines the router's state with past outcomes. In an idealized version, tasks fall into $J$ cells that share a record; ACRouter's embedding-keyed store~\citep{zhou2026agent} is a richer instance. Cells arrive independently with probabilities $\nu_j>0$. At the start of a round, each cell is uninformed ($U$) or holds an informative ($I$) or erroneous ($W$) record. After the round, each record expires with probability $\delta_j$, and the arriving cell receives an informative write with probability $\omega_j$ or an erroneous one with probability $\psi_j$, replacing its record; write types are independent of the past. Put $r_j=\omega_j+\psi_j>0$ and $\alpha_j=(1-\delta_j)(1-\nu_jr_j)<1$. Let $v_j^s$ and $c_j^s$ be the router's expected value and cost in cell $j$ and record state $s\in\{U,I,W\}$, including the record update. Records are measured against the uninformed router, which writes at the same cost but routes as if uninformed and has efficiency $\eta_U=\sum_j\nu_jv_j^U/\sum_j\nu_jc_j^U$, by the gains $g_j^s=(v_j^s-v_j^U)-\eta_U(c_j^s-c_j^U)$.

\begin{proposition}[Stationary memory gain]
\label{thm:memory}
Let $\chi_j=\nu_jr_j/(1-\alpha_j)$ be the stationary coverage of cell $j$ and $e_j=\psi_j/r_j$ its erroneous write fraction. The stationary efficiency $\eta_\infty$ satisfies $\eta_\infty\ge\eta_U$ if and only if
\begin{equation}
\sum_j\nu_j\chi_j\bigl[(1-e_j)g_j^I+e_jg_j^W\bigr]\ \ge\ 0 .
\label{eq:main-memory}
\end{equation}
If $g_j^I>0>g_j^W$, cell $j$ contributes positively if and only if $e_j<g_j^I/(g_j^I-g_j^W)$.
\end{proposition}
\begin{proof}
A record before round $n+1$ was written in round $n$ or survived it, so the probabilities of informative and erroneous records satisfy $i_{j,n+1}=\nu_j\omega_j+\alpha_ji_{j,n}$ and $w_{j,n+1}=\nu_j\psi_j+\alpha_jw_{j,n}$, with stationary values $i_j=\chi_j(1-e_j)$ and $w_j=\chi_je_j$. Conditioning on the arriving cell, the stationary surplus per round is $\E[V-\eta_UD]=\sum_j\nu_j\bigl[v_j^U-\eta_Uc_j^U+i_jg_j^I+w_jg_j^W\bigr]$, whose uninformed terms sum to zero by the definition of $\eta_U$. The result is the left side of Eq.~\eqref{eq:main-memory}, and it equals $\bar c_\infty(\eta_\infty-\eta_U)$ for the stationary mean cost $\bar c_\infty>0$. The cell condition follows from $(1-e_j)g_j^I+e_jg_j^W=g_j^I-e_j(g_j^I-g_j^W)$.
\end{proof}

\section{The Omni-Modal Trajectory Matrix: Details}
\label{app:matrix}

\subsection{Benchmarks, Metrics, and Candidates}
\label{app:benchmarks}
The 25 benchmarks form seven groups:
\begin{itemize}[leftmargin=1.2em,itemsep=1pt,topsep=2pt]
\item \emph{text QA}: MMLU-Pro~\citep{wang2024mmlupro}, MMLU-Redux~\citep{gema2025mmluredux}, SuperGPQA~\citep{du2025supergpqa}, GPQA~\citep{rein2024gpqa}, AA-LCR~\citep{aa2026evals}, AA-Omniscience~\citep{jackson2025omniscience}, and the text subset of HLE~\citep{phan2025hle};
\item \emph{image-text QA}: MMMU-Pro~\citep{yue2025mmmupro}, MMMU~\citep{yue2024mmmu}, and VLM-ExecRouterBench~\citep{yu2026scope};
\item \emph{image generation}: GenEval2~\citep{kamath2025geneval2}, Qwen-Image-Bench~\citep{li2026qwenimagebench}, LongT2I~\citep{yang2026longt2ibench}, OneIG~\citep{chang2025oneig}, and WISE~\citep{niu2025wise};
\item \emph{video generation}: VBench-2.0~\citep{zheng2025vbench2}; \emph{speech generation}: Seed-TTS English and Chinese~\citep{anastassiou2024seedtts};
\item \emph{code and workplace agents}: AA-Briefcase~\citep{aa2026evals}, AutomationBench~\citep{shepard2026automationbench}, GDP.pdf~\citep{garre2026gdppdf}, GDPval~\citep{patwardhan2025gdpval}, SciCode~\citep{tian2024scicode}, and Terminal-Bench~\citep{tbench2025};
\item \emph{robot manipulation}: RoboTwin~2.0~\citep{chen2025robotwin2}, 50 bimanual tasks in clean and randomized scenes.
\end{itemize}
The matrix holds 374,401 scored runs on 103,307 tasks from 452 candidate configurations, which differ in model, harness, and effort. Understanding benchmarks score a binary answer (AA-Omniscience uses plain accuracy). GenEval2 uses its official Soft-TIFA-GM score, Qwen-Image-Bench its official hierarchical total, LongT2I a learned alignment score, OneIG the mean of the sub-scores that apply to a single image, and WISE binary correctness. VBench-2.0 averages the dimensions that apply to each prompt, so it differs from the official total. Seed-TTS scores intelligibility as $1-\mathrm{WER}$ (English) or $1-\mathrm{CER}$ (Chinese), clipped to $[0,1]$. AA-Briefcase scores the fraction of passed checks, AutomationBench assertion credit, GDP.pdf whether every criterion passes, GDPval the rubric fraction, SciCode step pass on the original tests, Terminal-Bench the official verifier reward, and RoboTwin~2.0 binary episode success. Agent candidates run in their native harness at maximum effort; understanding candidates answer in one zero-shot request at high effort.

The candidates are: Claude Opus~5.5, GPT-6.1 Sol, DeepSeek V4.1 Flash, and, for long-context, knowledge, and agent tasks, GPT-6 Astra; the image generators FLUX~3, Seedream~5.0 Pro, MAI-Image~2.6, GPT Image~2.5, Qwen-Image~3 Pro, and Grok Imagine~2.0 (two further generators appear on a few GenEval2 prompts); the video generators MiniMax H3, Seedance~2.5, and Wan~3.0; the speech generators Seed Audio~1.0, Fish S2.1 Pro, and Gemini~3.8 Flash TTS; and the policies $\pi_{0.5}$~\citep{black2025pi05}, GigaBrain, Spatial Forcing~\citep{li2026spatialforcing}, X-VLA~\citep{zheng2026xvla}, X-WAM, PatchWAM, and ME-Dex, all co-trained on the 50 RoboTwin tasks. AA-LCR, GDP.pdf, and Terminal-Bench run every task three, five, and three times; each run is a separate input. Table~\ref{tab:app-headroom} gives the per-benchmark potential gain behind Table~\ref{tab:headroom}.

\begin{table}[t]
\centering\footnotesize\setlength{\tabcolsep}{4pt}
\caption{Per-benchmark potential gain on the test split (Table~\ref{tab:headroom}). The best fixed candidate in hindsight is named when it differs from the fixed candidate chosen on training inputs.}
\label{tab:app-headroom}
\begin{tabular}{@{}lrlrlrr@{}}
\toprule
Benchmark & Inputs & Fixed candidate & Fixed & Best fixed in hindsight & Oracle & Gap\\
\midrule
\multicolumn{7}{@{}l}{\emph{Text QA}}\\
\quad MMLU-Pro & 2,403 & Opus 5.5 & 92.55 & 92.55 & 95.21 & 2.66\\
\quad MMLU-Redux & 1,080 & Opus 5.5 & 96.67 & 96.67 & 98.06 & 1.39\\
\quad SuperGPQA & 5,300 & Opus 5.5 & 78.26 & 78.26 & 83.66 & 5.40\\
\quad GPQA (extended) & 106 & Opus 5.5 & 90.57 & 90.57 & 92.45 & 1.89\\
\quad AA-LCR & 42 & Opus 5.5 & 97.62 & 97.62 & 100.00 & 2.38\\
\quad AA-Omniscience & 120 & Opus 5.5 & 77.50 & 77.50 & 85.83 & 8.33\\
\quad HLE (text) & 431 & Opus 5.5 & 58.00 & 58.00 & 64.50 & 6.50\\
\multicolumn{7}{@{}l}{\emph{Image-text QA}}\\
\quad MMMU-Pro & 350 & Opus 5.5 & 86.86 & 86.86 & 92.86 & 6.00\\
\quad MMMU (dev/val) & 94 & Opus 5.5 & 81.91 & 81.91 & 84.04 & 2.13\\
\quad VLM-ExecRouterBench & 5,264 & Opus 5.5 & 85.28 & 85.28 & 87.77 & 2.49\\
\multicolumn{7}{@{}l}{\emph{Image generation}}\\
\quad GenEval2 & 160 & GPT Image 2.5 & 82.48 & 82.48 & 96.18 & 13.69\\
\quad Qwen-Image-Bench & 193 & GPT Image 2.5 & 64.09 & 64.09 & 69.83 & 5.73\\
\quad LongT2I & 543 & MAI-Image 2.6 & 86.14 & 86.14 & 86.89 & 0.75\\
\quad OneIG & 427 & FLUX 3 & 83.35 & 83.35 & 89.34 & 5.99\\
\quad WISE & 192 & Qwen-Image 3 & 92.19 & 92.71 (MAI-Image 2.6) & 98.96 & 6.77\\
\multicolumn{7}{@{}l}{\emph{Video generation}}\\
\quad VBench-2.0 & 158 & MiniMax H3 & 62.43 & 62.43 & 81.36 & 18.93\\
\multicolumn{7}{@{}l}{\emph{Speech generation}}\\
\quad Seed-TTS (en) & 214 & Fish S2.1 & 98.41 & 98.43 (Gemini 3.8 TTS) & 99.02 & 0.61\\
\quad Seed-TTS (zh) & 394 & Fish S2.1 & 98.73 & 98.73 & 99.45 & 0.72\\
\multicolumn{7}{@{}l}{\emph{Code and workplace agents}}\\
\quad AA-Briefcase & 1 & Opus 5.5 & 66.67 & 75.00 (GPT-6.1 Sol) & 75.00 & 8.33\\
\quad AutomationBench & 160 & Opus 5.5 & 92.75 & 92.75 & 96.53 & 3.78\\
\quad GDP.pdf & 100 & GPT-6.1 Sol & 21.00 & 23.00 (GPT-6 Astra) & 36.00 & 15.00\\
\quad GDPval & 43 & Opus 5.5 & 91.18 & 91.18 & 92.79 & 1.61\\
\quad Terminal-Bench & 39 & Opus 5.5 & 71.79 & 71.79 & 87.18 & 15.38\\
\multicolumn{7}{@{}l}{\emph{Robot manipulation}}\\
\quad RoboTwin 2.0 & 2,000 & ME-Dex & 70.45 & 70.45 & 88.85 & 18.40\\
\bottomrule
\end{tabular}
\end{table}

\subsection{Splits}
\label{app:splits}
The full set has 104,050 decisions over 25 benchmarks, split 70/10/20 with seed 42 by source group, so identical inputs and items from one source, such as the questions shared by MMMU-Pro and VLM-ExecRouterBench, never cross splits; existing research splits are kept, including RoboTwin's split by task family (35 training, 5 development, and 10 test tasks). This gives 72,351 training, 10,630 development, and 21,069 test records; the router studies of Section~\ref{sec:routers} use the first release of 18 benchmarks (Table~\ref{tab:panels}).

\subsection{Routers}
\label{app:routers}
Fable~5.1, a retired model, is excluded from every evaluation; routers trained with it play Opus~5.5 in its place.
\paragraph{Capability-tree router.} Leaves are defined per task and labeled on training inputs by GPT-6.1 Sol, which sees the input, the output requirement, and the leaf definitions, but no candidate outcomes. The classifier is Qwen3.5-4B with LoRA (rank 8), trained for three epochs on all 18 benchmarks at once with batch 16 and learning rates $10^{-4}$ (adapter) and $10^{-3}$ (head); the development split selects the epoch. The leaf table is fitted on training inputs by least squares over paired candidate differences on shared inputs.
\paragraph{Outcome and type routers on 11 benchmarks.} Both use Qwen3.5-4B with LoRA (rank 16, $\alpha=32$) on every linear layer of the language model, a final-token linear head, AdamW with learning rates $2\times10^{-5}$ (adapter) and $3\times10^{-4}$ (head), 32 tasks per update, and six epochs, with seeds 42, 13, 2026, and 7. The outcome router reads one (input, candidate profile) pair per forward pass; its loss is the cross-entropy against the normalized outcome plus four times the squared error of within-task score differences. The type router predicts one of 68 capability types and reads a global type-to-candidate table fitted on training inputs.
\paragraph{Robot router.} The outcome router uses Qwen3.5-4B with LoRA (rank 8), seven sigmoid heads, three epochs, batch 16, and learning rates $10^{-4}$ (adapter) and $10^{-3}$ (head); development routing success selects the epoch. The winner-label classifier shares the backbone, inputs, and data, with a seven-way softmax over the policy that succeeds, ties broken by measured inference time.
\paragraph{Image-routing suite.} Thirty-one router designs, among them RouterDC~\citep{chen2024routerdc}, SCOPE~\citep{yu2026scope}, IRT-based routers, $k$NN, MLP, and two-tower scorers, and Qwen3.5 label routers with 0.8B to 9B parameters, are trained on GenEval2 or Qwen-Image-Bench prompts and tested on the same or the other benchmark: 124 configurations, with three seeds for stochastic methods. Each is compared with the fixed generator by a paired bootstrap over prompts.
\paragraph{Untrained router.} Qwen3.5-4B, prompted with the input and the candidates, names the candidate to run.
\paragraph{Winner-label routers on 16 benchmarks.} Qwen3.5-4B is fully fine-tuned to emit the label of the best candidate, ties broken by cost, tokens, and a fixed order. The 12 recipes start from the instruction-tuned or a continually pretrained model, with or without fine-tuning on public routing data, and train per benchmark, on all benchmarks at once, or both in sequence.
Tables~\ref{tab:app-routers} and~\ref{tab:app-objectives} give per-benchmark results of the capability-tree router and of the 11-benchmark study.

\begin{table}[t]
\centering\footnotesize\setlength{\tabcolsep}{4pt}
\caption{Per-benchmark results of the capability-tree router (first block of Table~\ref{tab:routers}); one training seed. SciCode steps are aligned across models; 175 of its 187 inputs have one scored candidate.}
\label{tab:app-routers}
\begin{tabular}{@{}lrlrrr@{}}
\toprule
Benchmark & Inputs & Fixed candidate & Fixed & Router & Oracle\\
\midrule
AA-LCR & 42 & Opus 5.5 & 97.62 & 95.24 & 100.00\\
AA-Omniscience & 120 & Opus 5.5 & 77.50 & 77.50 & 85.83\\
HLE (text) & 426 & Opus 5.5 & 58.45 & 58.45 & 64.55\\
SciCode & 187 & Opus 5.5 & 70.05 & 70.05 & 70.59\\
Terminal-Bench & 36 & Opus 5.5 & 69.44 & 77.78 & 86.11\\
AutomationBench & 160 & Opus 5.5 & 92.75 & 92.75 & 96.53\\
GDP.pdf & 99 & GPT-6.1 Sol & 21.21 & 22.22 & 36.36\\
GDPval & 37 & Opus 5.5 & 91.84 & 91.84 & 93.62\\
GenEval2 & 162 & GPT Image 2.5 & 82.66 & 82.66 & 96.18\\
Qwen-Image-Bench & 197 & GPT Image 2.5 & 64.25 & 64.25 & 69.93\\
LongT2I & 552 & MAI-Image 2.6 & 86.13 & 86.12 & 86.89\\
OneIG & 450 & FLUX 3 & 83.20 & 84.30 & 89.20\\
WISE & 198 & Qwen-Image 3 & 92.42 & 91.92 & 98.99\\
VBench-2.0 & 161 & MiniMax H3 & 62.17 & 64.28 & 80.22\\
Seed-TTS (en) & 216 & Fish S2.1 & 98.42 & 98.42 & 99.02\\
Seed-TTS (zh) & 401 & Fish S2.1 & 98.74 & 98.74 & 99.45\\
\midrule
All 16 benchmarks & 3,444 & & 79.68 & 79.98 & 84.80\\
\bottomrule
\end{tabular}
\end{table}

\begin{table}[t]
\centering\footnotesize\setlength{\tabcolsep}{4pt}
\caption{Per-benchmark results of the 11-benchmark study (first block of Table~\ref{tab:objectives}), averaged over four training seeds; the fixed candidate is matched.}
\label{tab:app-objectives}
\begin{tabular}{@{}lrlrrrr@{}}
\toprule
Benchmark & Inputs & Fixed candidate & Fixed & Outcome router & Type router & Oracle\\
\midrule
AA-LCR & 42 & Opus 5.5 & 97.62 & 95.24 & 97.62 & 100.00\\
AA-Omniscience & 120 & Opus 5.5 & 77.50 & 77.29 & 71.88 & 85.83\\
HLE (text) & 431 & Opus 5.5 & 58.24 & 57.95 & 56.61 & 64.27\\
AA-Briefcase & 1 & GPT-6.1 Sol & 75.00 & 62.50 & 50.00 & 75.00\\
AutomationBench & 160 & Opus 5.5 & 92.75 & 92.70 & 92.54 & 96.53\\
GDP.pdf & 100 & GPT-6.1 Sol & 21.00 & 20.50 & 20.00 & 36.00\\
GDPval & 43 & Opus 5.5 & 90.52 & 90.17 & 90.31 & 92.20\\
SciCode & 15 & GPT-6.1 Sol & 93.33 & 93.33 & 86.67 & 93.33\\
Terminal-Bench & 39 & Opus 5.5 & 69.23 & 68.59 & 67.31 & 87.18\\
GenEval2 & 158 & GPT Image 2.5 & 82.56 & 82.18 & 82.56 & 96.13\\
Qwen-Image-Bench & 193 & GPT Image 2.5 & 64.09 & 63.59 & 64.09 & 69.83\\
\midrule
All 11 benchmarks & 1,302 & & 68.30 & 67.90 & 66.98 & 75.84\\
\bottomrule
\end{tabular}
\end{table}

\subsection{Price Routes in Image and Speech Generation}
\label{app:price-image}
Candidates are the generators scored on every input of a benchmark (Table~\ref{tab:app-price}). For each tolerance $\epsilon$, we take the most efficient price-path router that meets the floor on training prompts and evaluate it on test prompts with a paired bootstrap over prompts. With capability leaves as states, the fitted zero-loss route gains 0.05 points on training prompts but loses 0.33 $[-0.90,+0.25]$ on test prompts at $1.34\times$ efficiency. Prices per image are 208.8--211.3 m\$ for GPT Image~2.5, 60.0 for Grok Imagine~2.0, 47.7--48.0 for FLUX~3, 45.0 for Seedream~5.0, 40.0 for Qwen-Image~3, and 39.0--39.4 for MAI-Image~2.6; on GenEval2 only GPT Image~2.5, FLUX~3, and Qwen-Image~3 are scored on every prompt. In speech generation, the fixed generator, Fish S2.1 (0.9--1.1 m\$ per output), is the best and among the cheapest at every tolerance up to two points; Gemini~3.8 TTS costs 1.0--1.2 m\$ and Seed Audio~1.0 13.5--16.0 m\$.

\begin{table}[t]
\centering\footnotesize\setlength{\tabcolsep}{4pt}
\caption{Image price routes: test quality (0--100) and price (m\$ per image) of the fixed generator and of the half-point route of Table~\ref{tab:price}, fitted on training prompts.}
\label{tab:app-price}
\begin{tabular}{@{}lrlrrlrr@{}}
\toprule
& & \multicolumn{3}{c}{Fixed generator} & \multicolumn{3}{c}{Half-point route}\\
\cmidrule(lr){3-5}\cmidrule(l){6-8}
Benchmark & Test prompts & Candidate & Quality & m\$ & Candidate & Quality & m\$\\
\midrule
GenEval2 & 160 & GPT Image 2.5 & 82.48 & 208.8 & GPT Image 2.5 & 82.48 & 208.8\\
Qwen-Image-Bench & 193 & GPT Image 2.5 & 64.09 & 211.1 & Grok Imagine 2.0 & 61.16 & 60.0\\
LongT2I & 543 & MAI-Image 2.6 & 86.14 & 39.4 & MAI-Image 2.6 & 86.14 & 39.4\\
OneIG & 427 & FLUX 3 & 83.35 & 48.0 & Qwen-Image 3 & 83.02 & 40.0\\
WISE & 192 & Qwen-Image 3 & 92.19 & 40.0 & Qwen-Image 3 & 92.19 & 40.0\\
\bottomrule
\end{tabular}
\end{table}

\FloatBarrier
\section{Code and Text Testbeds: Details}
\label{app:experiments}

\subsection{Data}
\label{app:data}
CodeRouterBench~\citep{zhou2026agent} gives outcomes and token prices for eight models (claude-opus-4-6, claude-sonnet-4-6, gpt-5.4, Qwen3-Max, Kimi-K2.5, Qwen3.5-Plus, glm-5, and MiniMax-M2.7) on 9,999 coding tasks: 7,080 probing tasks (train and validation combined) and 2,919 test tasks. RouterBench~\citep{hu2024routerbench} gives the 0-shot matrix of 36,497 prompts and eleven models. Dollar costs are token-price estimates; sequential seconds add recorded primary and fallback latencies. Policy, portfolio, retry, and signal studies replay recorded outcomes. Verifier verdicts and the LLM forecast are fresh calls through one OpenAI-compatible relay, with recorded Opus outcomes as the fallback; the memory study is semi-synthetic. Scores are on the benchmarks' 0--1 scale, whereas the main text reports quality changes in points. Intervals are 95\% percentile task-bootstrap intervals, paired for policy comparisons, and efficiency is a ratio of averages.

\subsection{Price Routes}
\label{app:policy-details}
Table~\ref{tab:x9-oos} gives the CodeRouterBench policies of Section~\ref{sec:policy-choice}. With states, every dollar-optimal policy dispatches each task to a non-Opus model, for both partitions and every fitted floor, whereas the best fixed cascade gains 0.0558 at only $2.00\times$. On the 21-cell menu, the zero-loss policy has multiplier $\kappa=5.43$ and constrained optimum $\eta^\star_\epsilon=927.7$ score per dollar, so $\lambda_\epsilon=144.3=4.16\,\eta_B$. It maximizes $J_{\lambda_\epsilon}$, as Theorem~\ref{thm:optimal-routing}(ii) predicts, and is therefore also the quality maximizer at its own budget and, since $\kappa>0$, the cheapest policy that meets the floor. Hand-picked operating points land elsewhere: $\lambda=\eta_B$ gives $10.85\times$, and a budget equal to Opus's cost $2.16\times$. Refitting the policy inside 500 replicates that halve and bootstrap the probing tasks gives $25.72\times$ $[23.01,28.04]$ at $\Delta V=-0.0039$ $[-0.0178,+0.0099]$; choosing a safety margin by a plain probing bootstrap instead gives margins of 0.015--0.0175 and test compliance of only 0.47--0.96.

\begin{table}[t]
\centering\footnotesize\setlength{\tabcolsep}{3pt}
\caption{CodeRouterBench policies on the 2,919 test tasks. Dollar efficiency $\eta/\eta_B$, cost in m\$ per task, and $\Delta V=\E[V-Y_B]$, with paired 95\% intervals. Floors are fitted on probing; at $\epsilon\ge0.02$ they are slack and give the unconstrained policy. Fixed dispatch plays the best probing model, which is also the cheapest (Qwen3-Max); the per-task oracle, an in-sample ceiling, gives each task the candidate maximizing $V-\eta^\star C$ on its realized outcomes.}
\label{tab:x9-oos}
\begin{tabular}{@{}L{0.29\textwidth}L{0.23\textwidth}rL{0.29\textwidth}@{}}
\toprule
Policy & $\eta/\eta_B$ [CI] & Cost & $\Delta V$ [CI]\\
\midrule
21 cells, unconstrained & 28.187 [26.840, 29.659] & 0.395 & $-0.0201$ [$-0.0343$, $-0.0056$]\\
21 cells, $\epsilon=0$ & 25.516 [24.278, 26.903] & 0.452 & $-0.0044$ [$-0.0175$, $+0.0097$]\\
21 cells, $\epsilon=0.01$ & 27.612 [26.311, 29.072] & 0.409 & $-0.0140$ [$-0.0272$, $-0.0004$]\\
9 dimensions, unconstrained & 28.024 [26.713, 29.452] & 0.398 & $-0.0192$ [$-0.0332$, $-0.0047$]\\
9 dimensions, $\epsilon=0$ & 22.733 [21.584, 23.954] & 0.502 & $-0.0090$ [$-0.0227$, $+0.0060$]\\
Fixed Qwen3-Max dispatch & 20.097 [18.965, 21.297] & 0.525 & $-0.0416$ [$-0.0549$, $-0.0275$]\\
Best fixed cascade & 2.004 [1.881, 2.140] & 6.556 & $+0.0558$ [$+0.0461$, $+0.0660$]\\
ACRouter released map & 1.745 [1.666, 1.826] & 7.643 & $+0.0631$ [$+0.0528$, $+0.0731$]\\
Per-task oracle & 33.510 [32.065, 35.206] & 0.368 & $+0.0248$ [$+0.0117$, $+0.0376$]\\
Always-Opus & 1.000 & 11.655 & 0\\
\bottomrule
\end{tabular}
\end{table}

\paragraph{Certification.}
Each of 200 splits uses a random half of probing as the fitting sample $\mathcal A$ (3,540 tasks) and the other half as $\mathcal B$, fits the price path on $\mathcal A$, tests it on $\mathcal B$ at $\alpha=0.05$, and evaluates the selected router on the test set. The full probing path has 38 routers, against the bound of 442. At $\epsilon=0$, the certified router has median efficiency $15.17\times$ with Hoeffding $p$-values, $18.83\times$ with empirical-Bernstein ones, and $19.38\times$ with empirical-Bernstein ones on the mixture-refined path (Appendix~\ref{app:proof-certified-floor}); Bonferroni selection over the same path gives $13.29\times$ and $15.17\times$. In 200 nested 70/30 resplits of all 9,999 tasks, the refined procedure holds the floor in 199 at a median $20.56\times$. The guarantee is an aggregate one: the certified router loses quality in a median 8 of 21 test cells. Certification needs every candidate on the path and the reference on $\mathcal B$, at \$0.0158 per task against \$0.0124 for Opus alone.

\paragraph{States without oracle labels, and learned routers.}
Prompts are rebuilt with the benchmark loaders (9,996 of 9,999 tasks). Predicted-state policies refit candidate values on five-fold out-of-fold predictions, so fitting and deployment see the same noisy state; a second variant first strips every line that occurs in at least 20\% of one source's prompts, by one rule learned on probing prompts (19\% of test-prompt characters), so that source wrappers cannot reveal the cell. Learned quality--cost routers predict each model's score and log cost from TF-IDF features by ridge regression and play $\max_m\hat v_m-\lambda\hat c_m$, directly, as a cascade, or under a mean-cost budget; the RouteLLM-style router~\citep{ong2024routellm} classifies whether Opus beats Qwen3-Max. The FrugalGPT-style cascade~\citep{chen2023frugalgpt} learns a TF-IDF scorer per model and searches all 400 chains of at most three models, and the AutoMix-style cascade~\citep{aggarwal2024automix} accepts the small model's answer when a meta-verifier accepts the recorded grok-4.6 verdict. Operating points are chosen on probing (Table~\ref{tab:policy-robustness}). Refitting the stripping rule, features, and classifier inside 200 replicates gives predicted cells a median $24.83\times$ at $\Delta V=-0.0026$: routing from prompt text works within the distribution.

\begin{table}[t]
\centering\footnotesize\setlength{\tabcolsep}{3pt}
\caption{CodeRouterBench states without oracle labels, and learned routers. $r=\eta/\eta_B$; cost in m\$ per task; compliance is the share of test resamples with $\Delta V\ge0$.}
\label{tab:policy-robustness}
\begin{adjustbox}{max width=\linewidth}
\begin{tabular}{@{}llrrrr@{}}
\toprule
States or router & Operating point & $r$ [95\% CI] & Cost & $\Delta V$ [95\% CI] & Compl.\\
\midrule
True 21 cells & $\epsilon=0$ & 25.52 [24.26, 26.77] & 0.452 & $-0.0044$ [$-0.0177$, $+0.0077$] & 0.22\\
Predicted 21 cells & $\epsilon=0$ & 25.32 [24.07, 26.63] & 0.453 & $-0.0070$ [$-0.0203$, $+0.0058$] & 0.13\\
Pooled (no state) & $\epsilon=0$ & 3.62 [3.41, 3.85] & 3.232 & $+0.0021$ [$-0.0085$, $+0.0135$] & 0.63\\
\multicolumn{6}{@{}l}{\emph{Source wrappers removed from every prompt by one rule learned on probing prompts}}\\
Predicted 21 cells & $\epsilon=0$ & 24.92 [23.65, 26.28] & 0.464 & $-0.0032$ [$-0.0163$, $+0.0096$] & 0.28\\
\midrule
Quality-cost, direct & probing $\epsilon\approx0$ & 24.17 [22.83, 25.49] & 0.482 & $-0.0003$ [$-0.0140$, $+0.0125$] & --\\
Quality-cost, cascade & probing $\epsilon\approx0$ & 22.67 [20.48, 24.72] & 0.512 & $-0.0016$ [$-0.0142$, $+0.0114$] & --\\
Quality-cost, budget & probing $\epsilon\approx0$ & 23.57 [22.21, 24.90] & 0.497 & $+0.0026$ [$-0.0107$, $+0.0157$] & --\\
$k$NN-20, direct & probing $\epsilon\approx0$ & 23.58 [22.28, 24.89] & 0.500 & $+0.0050$ [$-0.0080$, $+0.0175$] & --\\
RouteLLM-style & probing $\epsilon\approx0$ & 6.36 [5.82, 6.90] & 1.861 & $+0.0064$ [$-0.0064$, $+0.0193$] & --\\
FrugalGPT-style cascade & probing $\epsilon\approx0$ & 11.76 [11.02, 12.51] & 1.000 & $+0.0039$ [$-0.0093$, $+0.0171$] & 0.71\\
\midrule
\multicolumn{6}{@{}l}{\emph{334 runnable test tasks with recorded judge verdicts; always-Opus scores 0.737 at 5.413 m\$}}\\
AutoMix-style, judge free & probing $\epsilon\approx0$ & 1.64 [1.47, 1.89] & 3.289 & $-0.0030$ [$-0.0240$, $+0.0150$] & 0.43\\
AutoMix-style, judge at list price & probing $\epsilon\approx0$ & 0.28 [0.26, 0.31] & 19.125 & $+0.0000$ [$-0.0180$, $+0.0180$] & 0.55\\
\bottomrule
\end{tabular}
\end{adjustbox}
\end{table}

\paragraph{A stronger yardstick.}
CellBest routes each cell to its best probing model; the Opus-referenced zero-loss policy falls 0.087 below it. Refitted against CellBest, the zero-loss policy reaches $3.18\times$ $[2.94,3.40]$ CellBest's efficiency at $\Delta V'=-0.0039$ $[-0.0108,+0.0027]$; a floor certifies quality only relative to the reference it names.

\paragraph{RouterBench.}
The reference, gpt-4-1106-preview, scores 0.781 at 3.29\,m\$ per test prompt. The menu has 21 candidates: direct dispatch to each of the eleven models, and each other model followed by the reference when its released score is below $\theta=1$. States are the eight source families, the 86 datasets (57 MMLU subsets), datasets predicted from the prompt (92.0\% test accuracy), TF-IDF $k$-means clusters, or one pooled state (Table~\ref{tab:routerbench-policy}). With family states, the best router that avoids the reference scores 0.097 below it, and the admission price $5.85\,\eta_B$ exceeds the zero-loss floor price $1.60\,\eta_B$, so the zero-loss router calls the reference: it sends 68\% of prompts to cascades, and the reference runs on 21\%. Certified on half of the fitting prompts, the zero-loss floor holds on the test prompts in all 200 halvings at a median $6.70\times$; uncertified fitted floors hold on 4 of 11 splits with family states and on none with dataset states. At matched cost, learned quality--cost cascades lie 0.014--0.022 above the 8-family frontier and tie with the 86-dataset one, so prompt-level prediction adds value where states are coarse.

\begin{table}[t]
\centering\footnotesize\setlength{\tabcolsep}{3pt}
\caption{Fixed-menu policies on 10,955 RouterBench test prompts at $\theta=1$. $r=\eta/\eta_B$ against gpt-4-1106-preview; cost in m\$ per prompt; compliance is the share of resamples with $\Delta V\ge0$.}
\label{tab:routerbench-policy}
\begin{adjustbox}{max width=\linewidth}
\begin{tabular}{@{}llrrrr@{}}
\toprule
States or router & Operating point & $r$ [95\% CI] & Cost & $\Delta V$ [95\% CI] & Compl.\\
\midrule
Best fixed cascade (Yi-34B, then reference) & probing & 1.38 [1.37, 1.39] & 2.515 & $+0.0425$ [$+0.0390$, $+0.0461$] & 1.00\\
8 families & unconstrained & 32.11 [31.37, 32.81] & 0.052 & $-0.3817$ [$-0.3910$, $-0.3722$] & 0.00\\
8 families & $\epsilon=0$ & 7.10 [6.91, 7.33] & 0.461 & $-0.0040$ [$-0.0084$, $+0.0008$] & 0.05\\
86 datasets & $\epsilon=0$ & 8.26 [8.01, 8.51] & 0.395 & $-0.0073$ [$-0.0125$, $-0.0024$] & 0.00\\
Predicted 86 datasets & $\epsilon=0$ & 7.85 [7.55, 8.16] & 0.416 & $-0.0057$ [$-0.0112$, $-0.0009$] & 0.01\\
TF-IDF $k$-means 20 & $\epsilon=0$ & 8.10 [7.82, 8.39] & 0.405 & $-0.0024$ [$-0.0079$, $+0.0026$] & 0.17\\
Pooled (no state) & $\epsilon=0$ & 1.71 [1.70, 1.73] & 1.910 & $-0.0035$ [$-0.0073$, $+0.0008$] & 0.06\\
\midrule
Quality-cost, cascade & price, fit $\epsilon\approx0$ & 8.10 [7.89, 8.32] & 0.405 & $-0.0025$ [$-0.0080$, $+0.0029$] & 0.18\\
Quality-cost, cascade & budget, fit $\epsilon\approx0$ & 7.07 [6.88, 7.27] & 0.474 & $+0.0152$ [$+0.0104$, $+0.0199$] & 1.00\\
$k$NN-20, cascade & price, fit $\epsilon\approx0$ & 6.71 [6.50, 6.94] & 0.492 & $+0.0026$ [$-0.0024$, $+0.0076$] & 0.83\\
\midrule
Per-task oracle (in-sample) & $\epsilon=0$ & 45.29 [44.68, 45.87] & 0.073 & $0.0000$ [$-0.0073$, $+0.0072$] & 0.48\\
\bottomrule
\end{tabular}
\end{adjustbox}
\end{table}

\subsection{Portfolio Selection}
\label{app:portfolio-details}
Candidates are fitted on probing and evaluated on test tasks. CodeRouterBench adds one of seven candidates, each with at least 20 passes and 20 failures, to portfolios of up to three models; RouterBench adds ten candidates to 21 initial portfolios over ten fixed 70/30 splits. The selecting verifier is an oracle that commits to the best passing member, else calls the reference, and every member call is charged. Paired selection picks the candidate with the highest fitted $\hat\eta(S^+)$ (Eq.~\eqref{eq:main-paired}); the controls rank residual coverage $N_m(S)$ (Eq.~\eqref{eq:main-residual}), an independence proxy that replaces $r_m(S)$ by $(1-p(S))\mathbb P(E_m)$, the saving score $r_m(S)b-d_m$, or marginal success (Table~\ref{tab:paired-candidate-selection}). Residual coverage evaluates the quality term at the current portfolio's conditional means, and those means shift: adding a model lowers the miss-conditional value in 97\% of CodeRouterBench comparisons, which Eq.~\eqref{eq:portfolio-shift-correction} restores. Adding Qwen3.5-Plus to the pooled Qwen3-Max portfolio, for example, is predicted to lose 1.4 score per dollar but gains $+5.0$ $[3.5,6.6]$, because the fallback cost of the new failure set falls by 14\%. Cluster-bootstrap intervals for the differences in best-candidate frequency, resampling CodeRouterBench dimensions or RouterBench source families, all include zero. Over ten CodeRouterBench resplits, mean relative regret at $\theta=1$ and $0.5$ is 0.04\% and 0.06\% for paired selection, 0.29\% and 0.05\% for residual coverage, and 13.1\% and 17.0\% for marginal success.

\begin{table}[t]
\centering\footnotesize\setlength{\tabcolsep}{3pt}
\caption{Best-candidate selection: the share of comparisons (one portfolio with several candidate additions) in which a rule's fitted choice has the highest test efficiency. Counts sum over splits, which share tasks. Paired ranks $G_m(S)$ of Eq.~\eqref{eq:main-paired}; residual ranks $N_m(S)$; Indep.\ assumes independent failures; saving ranks $r_mb-d_m$.}
\label{tab:paired-candidate-selection}
\begin{tabular}{@{}lrrrrrrr@{}}
\toprule
Dataset, threshold & Splits & Count & Paired & Residual & Indep. & Saving & Marginal\\
\midrule
CodeRouterBench, $\theta=1$ & 1 & 139 & \textbf{97.1\%} & 86.3\% & 86.3\% & 95.7\% & 46.8\%\\
CodeRouterBench, $\theta=0.5$ & 1 & 194 & \textbf{96.4\%} & 94.3\% & 91.8\% & 93.3\% & 40.7\%\\
RouterBench, $\theta=0.5$ & 10 & 5,539 & \textbf{63.9\%} & 61.9\% & 35.9\% & 63.5\% & 14.6\%\\
RouterBench, $\theta=1$ & 10 & 212 & \textbf{64.6\%} & 53.8\% & 36.3\% & 48.1\% & 0.0\%\\
\bottomrule
\end{tabular}
\end{table}

\subsection{Verifiers and Overheads}
\label{app:verifier-details}
On recorded responses of three primaries to 500 probing and 334 runnable test tasks, we compare a static \emph{Compiler} gate, which compiles the extracted answer and the official harness to bytecode and checks the declared entry point without executing code; visible examples; hidden tests, an oracle-like control; the released-score oracle; and a grok-4.6 judge on two primaries, charged at the primary's blended token price or at its list price of \$2 and \$6 per million input and output tokens (Table~\ref{tab:x8-verifiers}). Holding outputs and the threshold fixed, the judge improves efficiency exactly when its mean charge is below $\bar V_\theta/\eta_B-\bar D_{0,\theta}$, where $\bar V_\theta$ is the mean delivered score and $\bar D_{0,\theta}$ the mean cost before the judge. For Qwen3-Max at $\theta=0.5$ this break-even charge is \$0.00330 per task $[\$0.00285,\$0.00379]$; at $\theta=1$ it is negative for both primaries, so even a free judge does not help. Accepting every Qwen3-Max output reaches $19.65\times$ but loses $Q=-0.063$ $[-0.108,-0.024]$, which the floor excludes.

Router inference is timed on one Apple M4 core and priced at AWS Lambda's x86 list price for a 2\,GB function (\$$3.33\times10^{-5}$ per second, us-east-1, September 2026). Per task, the TF-IDF router takes 1.9\,ms (0.06\% of a Qwen3-Max call), a hidden-test sandbox run 0.15\,s and \$$5.2\times10^{-6}$ (2.0\%), the grok-4.6 judge 39.0\,s and \$0.0159 at list price ($63\times$), and the DeepSeek-v4-pro forecast 1.89\,s and \$0.000944 ($2.4\times$); charging router overhead leaves every replay comparison unchanged at the reported precision. Opus takes 3.93\,s per task against 6.23\,s for Qwen3-Max, so in sequential seconds direct dispatch reaches only $0.58\times$ (Qwen3-Max) and $0.50\times$ (Kimi-K2.5) Opus's efficiency, verified cascades $0.40$--$0.54\times$, and the judge's 39-second call lowers the ratio to $0.08\times$.

\begin{table}[t]
\centering\small
\caption{Verifier choice changes the cascade's efficiency and quality. $r=\eta/\eta_B$; $Q$ is mean score change. Qwen3-Max primary; the judge is charged at the primary's token price.}
\label{tab:x8-verifiers}
\begin{tabular}{@{}lcccc@{}}
\toprule
Verifier & AUROC & $r(\theta=0.5)$ [95\% CI] & $r(\theta=1)$ [95\% CI] & $Q(\theta=0.5)$\\
\midrule
Compiler (static gate) & 0.766 & 4.09 [3.23, 5.45] & 4.09 [3.23, 5.45] & $-0.024$\\
Visible examples & 0.537 & 1.98 [1.72, 2.35] & 1.06 [1.03, 1.10] & $-0.036$\\
Hidden tests & 0.994 & 2.87 [2.38, 3.53] & 2.37 [2.03, 2.80] & $+0.024$\\
LLM judge & 0.824 & 1.09 [1.00, 1.21] & 0.64 [0.61, 0.66] & $-0.012$\\
Released-score oracle & 1.000 & 2.46 [2.09, 2.93] & 2.46 [2.09, 2.93] & $+0.045$\\
\bottomrule
\end{tabular}
\end{table}

\subsection{Routing Signals}
\label{app:signal-details}
The released CodeRouterBench routers expose their chosen model, which we use as a binary signal $Z$ (chose Qwen3-Max), with posteriors cross-fitted over ten half-splits of the test set; capture is the gross saving divided by the perfect-information bound of Proposition~\ref{prop:voi} (Table~\ref{tab:x12a-signals}). Classical routers have negative native capture because their $Z=0$ tasks still satisfy $\mathbb P(H=1\mid Z=0)>a/b=0.043$, so withholding the attempt forgoes profitable ones. For the LLM forecast, DeepSeek-v4-pro reads the prompt and predicts whether Qwen3-Max and Kimi-K2.5 pass all hidden tests on 600 probing and 600 test tasks; its calibrated posterior reaches test AUROC 0.541 and 0.618 and a capture of $-0.037$ $[-0.225,0.077]$, and at list price it costs more than the attempt it could save in 22 of 24 states.

\begin{table}[t]
\centering\footnotesize\setlength{\tabcolsep}{3.5pt}
\caption{Cross-fitted capture of the nine released CodeRouterBench routers as signals $Z$ (chose Qwen3-Max) for Qwen3-Max at $\theta=1$, with no signal charge. Native use attempts on $Z=1$, threshold use when $s_Z>a/b$; the last column applies the threshold rule given the benchmark dimension and divides the aggregated saving by the aggregated bound.}
\label{tab:x12a-signals}
\begin{tabular}{@{}lccc@{}}
\toprule
Signal source & Native [95\% CI] & Threshold [95\% CI] & Within dimension\\
\midrule
Fine-tuned Qwen 27B & 0.555 [0.490, 0.614] & 0.555 [0.490, 0.614] & 0.062\\
Fine-tuned Qwen 9B & 0.546 [0.475, 0.608] & 0.546 [0.475, 0.608] & 0.078\\
Fine-tuned Qwen 2B & 0.466 [0.397, 0.531] & 0.417 [0.355, 0.479] & $-0.006$\\
Fine-tuned Qwen 0.8B & 0.422 [0.348, 0.491] & 0.082 [0.066, 0.098] & $-0.008$\\
Logistic regression & $-1.218$ [$-1.590$, $-0.869$] & 0 & $-0.002$\\
TF-IDF MLP & $-1.736$ [$-2.178$, $-1.342$] & 0 & $-0.012$\\
RouteLLM MF & $-1.033$ [$-1.364$, $-0.723$] & 0 & $-0.013$\\
RouteLLM SW & $-1.284$ [$-1.685$, $-0.953$] & 0 & $+0.005$\\
DimensionBest & $-1.468$ [$-1.847$, $-1.104$] & 0 & 0\\
\bottomrule
\end{tabular}
\end{table}

\subsection{Router Memory}
\label{app:memory-details}
This semi-synthetic study combines real outputs and verifier judgments with constructed arrivals and retention: 2,044 probing and 867 test tasks in 12 cells. With probability 0.25 the writer explores a uniformly sampled model, and it writes the model's identity to the arriving cell when verification accepts; the newest record serves later tasks in that cell, and uncovered cells use the prior, always-Opus or the probing-fitted DimensionBest, which sends each dimension to its best model. Writers use ACRouter's doctest and syntax tools, locally executed hidden tests, or a DeepSeek-v4-pro judge; we report hidden-test writes. A write is erroneous when the accepted output is incorrect under the released score; against that truth, hidden-test writes have TPR/FPR 0.960/0.205. Fifty streams of 60,000 rounds are measured over their last 10,000 rounds. With no expiration, the store reaches 135.8 score per dollar but scores 0.034 below always-Opus and 0.116 below DimensionBest.

\FloatBarrier
\section{Limitations}
\label{sec:discussion}

The theory takes the law of states as given, so it does not cover states that depend on the router's own earlier choices, such as SciCode steps that contain a model's earlier code, and certification needs the reference's outcome on every certification input. Moreover, step-level routing for long-horizon agentic tasks is more complex and deserves deeper exploration.

\end{document}